\documentclass[journal]{IEEEtran}
\usepackage{graphicx}
\usepackage{amsmath}
\usepackage{amssymb}
\usepackage{diagbox}             
\usepackage{multirow}
\usepackage{booktabs} 
\usepackage{times}
\usepackage{epsfig}
\usepackage[pagebackref=true,breaklinks=true,letterpaper=true,colorlinks,bookmarks=false]{hyperref}

\ifCLASSINFOpdf
\else
\fi
\begin{document}
%
\title{SSAGCN: Social Soft Attention Graph Convolution Network for Pedestrian Trajectory Prediction}
%
%
%


\author{Pei~Lv,
	Wentong~Wang,
	Yunxin~Wang,
	Yuzhen~Zhang,
	Mingliang~Xu ~\IEEEmembership{Member,~IEEE},
	and~Changsheng~Xu,~\IEEEmembership{Fellow,~IEEE}
	\thanks{Pei Lv, Yuzhen Zhang and Mingliang Xu are with the School of Computer and Artificial Intelligence, Zhengzhou University, Zhengzhou 450001, China. E-mail: {ielvpei, iexumingliang}@zzu.edu.cn, zyzzhang@gs.zzu.edu.cn}
	\thanks{Wentong Wang and Yunxin Wang are with Henan Institute of Advanced Technology, Zhengzhou University, Zhengzhou 450003, China. E-mail: {wangwentong, wangyunxin}@gs.zzu.edu.cn}
	\thanks{Changsheng Xu are with the NLPR, Institute of Automation, Chinese Academy of Sciences, Beijing 100864, China, and also with the University of Chinese Academy of Sciences, Beijing 100049, China. E-mail: changsheng.xu@ia.ac.cn}
}

%
%

\markboth{Journal of \LaTeX\ Class Files,~Vol.~14, No.~8, August~2015}%
{Shell \MakeLowercase{\textit{et al.}}: Bare Demo of IEEEtran.cls for IEEE Journals}
%



\maketitle

\begin{abstract}
Pedestrian trajectory prediction is an important technique of autonomous driving, which has become a research hot-spot in recent years. In order to accurately predict the reasonable future trajectory of pedestrians, it is inevitable to consider social interactions among pedestrians and the influence of surrounding scene simultaneously. Previous methods mainly rely on the position relationship of pedestrians to model social interaction, which is obviously not enough to represent the complex cases in real situations. In addition, most of existing  work usually introduce the scene interaction module as an independent branch and embed the social interaction features in the process of trajectory generation, rather than simultaneously carrying out the social interaction and scene interaction, which may undermine  the rationality of trajectory prediction. In this paper, we propose one new prediction model named Social Soft Attention Graph Convolution Network (SSAGCN) which aims to simultaneously handle social interactions among pedestrians and scene interactions between pedestrians and environments. In detail, when modeling social interaction, we propose a new \emph{social soft attention function}, which fully considers various interaction factors among pedestrians. And it can distinguish the influence of pedestrians around the agent based on different factors under various situations. For the physical interaction, we propose one new \emph{sequential scene sharing mechanism}. The influence of the scene on one agent at each moment can be shared with other neighbors through social soft attention, therefore the influence of the scene is expanded both in spatial and temporal dimension. With the help of these improvements, we successfully obtain socially and physically acceptable predicted trajectories. The experiments on public available datasets prove the effectiveness of SSAGCN and have achieved state-of-the-art results.
\end{abstract}

\begin{IEEEkeywords}
Graph convolutional network, Social soft attention, Social and scene interactions, Trajectory prediction.
\end{IEEEkeywords}

%
\IEEEpeerreviewmaketitle

\section{Introduction}
%
%
%
%
\IEEEPARstart{P}{redicting} the future trajectory of pedestrians in video is one significant task in autonomous driving~\cite{A1,A2,A3,A4,A5,A6}, which attracts plenty of scientists and engineers in different research fields nowadays. The successful prediction of the future trajectory of pedestrians is closely related to the accurate modeling of influence factors of the surrounding pedestrians and the physical scene.
\begin{figure}[t]
	\begin{center}
		\includegraphics[width=1.0\linewidth]{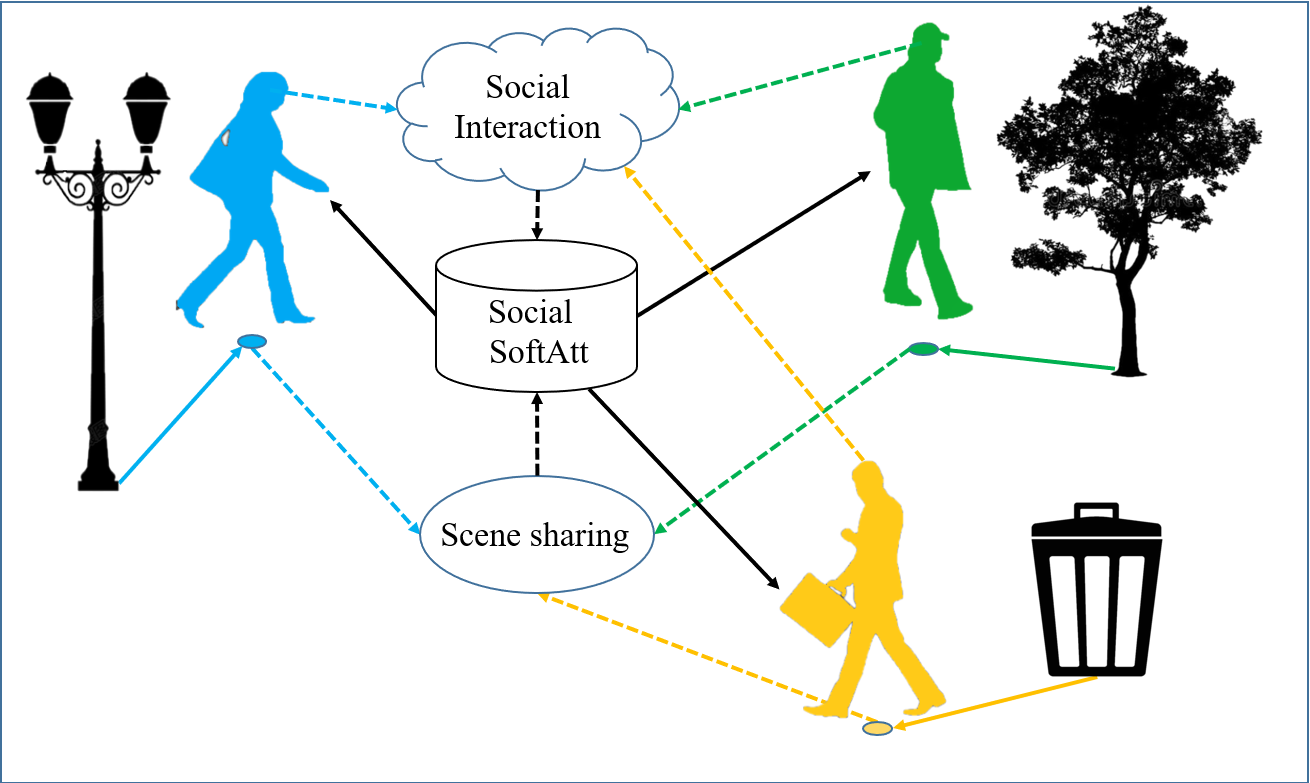}
	\end{center}
	\caption{Illustration of the interactions for pedestrians. The yellow, green and blue solid lines represent the process of perceiving the surrounding environment by different pedestrians. The yellow, green and blue dotted lines represent the process in which the impact of the scene is shared through the social interaction of different pedestrians. The black dotted line and black solid line represent the process of social interaction, scene attention sharing and feedback among pedestrians.} 
	\label{fig1}
\end{figure}
\par 
Early works aggregate the hidden state of the recurrent neural network to model pedestrian interactions according to the location information~\cite{B1, B2,B3,B4,B5,B6}. Afterwards the attention mechanism~\cite{C8} is introduced to calculate the degree of the influence of different pedestrians to other agents~\cite{B8, B9, B10}. 
Recently, graph structure is involved to represent the topological relationship of pedestrians~\cite{B15,A16,C1,E9}, and achieved competitive performance. 
However, the real situation is too  complex to be modeled by simple graph structure, where the social factors affecting the future trajectory of agents include relative speed, position, perspective, etc. People tend to be influenced by their neighbors in the field of view, and the effect will increase with larger relative speed and closer distance. Unfortunately, most of previous methods mainly take into account the relative distance and lack of comprehensive consideration on aforementioned factors.
 
\par In another aspect, people have realized the important influence of the scene to the problem of trajectory prediction~\cite{C6,C7,B11,B12,B13,B8}. Most of them try to take the original scene image as the input and utilize convolution neural network to extract corresponding scene features. These features are then used to calculate the impact of the scene on pedestrians. For example, Sophie~\cite{B8} and Social-BiGAT~\cite{B20} used fixed frames to calculate physical attention and considered social interaction and scene interaction separately. However, there are two obvious shortcomings in this manner. First, the physical scene usually plays an important role in the whole process of pedestrian movement, not just in fixed period of time. Second, the interactions among pedestrians usually occur simultaneously with that between pedestrians and scene. Above observations motivate us to fully consider the continuous influence of physical scene in our method, and share the impact of scene on the agents with the neighbors in time. 

\par In view of the excellent performance of graph-based methods~\cite{C1,B15}, we continue this way in our work. Compared with previous work, we  further analyze the social factors that may influence the trajectory of pedestrians comprehensively. For example, pedestrians who meet from opposite directions will have a greater impact on each other, and the larger the relative speed, the greater the impact. Due to the limited field of view, pedestrians who leave in opposite directions have less influence on each other, even if they are very close. Based on above practical laws, we propose a new \textbf{\textit{social soft attention function}} to cover these situations, and use it to construct a weighted adjacency matrix to represent different degrees of influence among pedestrian nodes in the graph. In addition, we also propose a \textbf{\textit{sequential scene attention sharing mechanism}}. By calculating the influence of scene in continuous moments, the effects of scene are firstly extended in the time domain, and then shared among pedestrians through social interactions. 

\par Moreover, our model takes into account various social interactive factors and physical scene information are embed into the unified graph node as never before. The GCN~\cite{B18} is utilized to process the node feature aggregation on the constructed graph, and the temporal convolutional Neural Networks(TCNs)~\cite{E15} are used to generate the predicted trajectory. The contributions can be summarized as follows:
\begin{itemize}
	\item We propose a new trajectory prediction model, SSAGCN, which can deal with more comprehensive social interactions among pedestrians, and also fully explores the sequential scene attention mechanism uncovered by previous work. Through comprehensively considering the impact of the scene on pedestrians and the interaction among pedestrians, the performance of SSAGCN on public datasets has reached the state-of-the-art results. 
	\item We propose a social soft attention function to calculate the degree of influence among pedestrian nodes in the spatial-temporal graph. This function fully considers the influence of relative speed, position and perspective factors on future trajectories of pedestrians.
	\item We propose a sequential scene attention sharing mechanism, which takes into account the physical effects of actual environment as part of graph nodes and shares these effects through the unified social propagation. The semantic information of graph nodes is largely enriched so that the proposed model can produce more physically acceptable prediction trajectories.
\end{itemize}

 
\section{Related Work}

With the rapid development of deep learning and extensive attention paid to autonomous driving technology, the pedestrian trajectory prediction has also been greatly promoted. Since there are huge amount of related work in this research area, we mainly focus on the following aspects and demonstrate them in details.
\subsection{Social interaction modeling among pedestrians}

 Researchers have noticed from the very beginning that it is inevitable to consider the influence of the surrounding neighbors on one agent to predict its movement in the future. Helbing~\emph{et al}.~\cite{A12} firstly proposed the concept of social force, which used the repulsive force and attraction between pedestrians to model social interaction. To calculate the impact of motionless crowds on pedestrian movement, Yi~\emph{et al}.~\cite{A13} introduced personality attributes to classify pedestrians into different categories. Social-LSTM~\cite{B4} was proposed to integrate the original vanilla LSTM with time-step pooling mechanism, and this model successfully simulated the social interaction of pedestrians. Since the importance of social influence is mainly determined by the distance between pedestrians and their neighbors, Xue~\emph{et al}.~\cite{B12} proposed to construct a circular occupancy map for each pedestrian to capture the impact of other pedestrians. Zhang~\emph{et al}.~\cite{B6} proposed a weighted algorithm similar to the attention mechanism, to extract social influence information from neighbors' current behavioral intentions to establish an interaction model. Sadeghian~\emph{et al}.~\cite{B8} proposed one social attention mechanism, which models the interaction between pedestrians by calculating the attention of the LSTM hidden state of nearby pedestrians. In recent years, some work~\cite{B15,C1} began to use graph structures to model social interaction. Huang~\emph{et al}.~\cite{B15}  proposed STGAT, which combined graph network and attention mechanism based on distance to share information among different pedestrians and assigned different weights to various nodes.  Social-STGCNN~\cite{C1} used graph convolutional network with more rich spatial-temporal information to encode the interactions among pedestrians.

\par Above methods have achieved satisfactory prediction results through modeling the social interactions among pedestrians. However, they did not  fully consider various influence factors mentioned before. In our method, we propose a new social soft attention function for graph convolution, which  considers the influence of position, relative speed and perspective factors comprehensively, to model the complex social interactions by calculating the degree of interaction between pedestrian nodes in graph structure.

\subsection{Scene interaction modeling between pedestrians and environment}

The scene plays a vital role in the trajectory prediction task. Recently, many methods~\cite{A9,A8,B8,A15,B9} focused on how to extract context features from the scene to constrain the movement of agents. Kitani~\emph{et al}.~\cite{E22} used the hidden variable Markov decision process to model the interaction between human and the scene and infer the passable area for pedestrians. Sadeghian~\emph{et al}.~\cite{A8} considered the dependencies between the historical trajectories of agents and the spatial navigation environment, and then presented an interpretable trajectory prediction framework. Similar to above methods, Sadeghian~\emph{et al}.~\cite{B8} further proposed the Sophie framework, which simultaneously modeled physical and social interactions by leveraging the history trajectory of agents and the scene context features. Moreover, some approaches~\cite{A15,B9} processed the original images of the scene into the form of binary segmentation that is to
distinguish the roads and obstacles of the scene. Haddad~\emph{et al}.~\cite{B9} took into account the interaction with static physical objects and dynamic pedestrians in the scene. They presented a new spatial-temporal graph based on Long Short-Term Memory (LSTM) network, which could avoid the potential collision in crowded environments efficiently.

In our work, we use physical scene attention mechanism to calculate the continuous effects of scenes on agents in sequential frames rather than specific ones. Moreover, in order to consider the influence of scene context and social interaction at the same time, we share the influence of scene at each moment with each adjacent node in the social interaction graph. Compared with previous methods, our approach expands the influence of the physical scene both in time and space.

\subsection{Graph neural network}

Graph neural networks (GNNs)~\cite{E1} are powerful neural network architecture for machine learning on graphs. Through combining GNNs and convolutional neural network, the graph convolutional network (GCN)~\cite{E2}, which allows for assigning various weights to different neighbors according to distances, has been widely used for various vision tasks. Recent years, Gao~\emph{et al}.~\cite{E3} used GCN to handle vision tracking tasks. Sun~\emph{et al}. ~\cite{B21} used GCN to process social interaction in video trajectory prediction tasks. Yan~\emph{et al}.~\cite{A17} utilized an improved model, STGCNN, to recognize human interactions. Abduallah~\emph{et al}~\cite{C1} formulate a kernel function to attach social attributes to STGCNN to predict pedestrian trajectory. However, these works solidify the degree of influence into the same value when modeling the influence of any two pedestrians on each other with graph structure, which is unreasonable in practical situations. In many cases the two pedestrians do not have exactly the same influence on each other. The emergence of GAT~\cite{E14} which brings attention to graph neural networks solves this problem well. Subsequently, both STGAT~\cite{B15} and Social-Bigat~\cite{A16} extended GAT in trajectory prediction. Shi~\emph{et al}~\cite{E16} used a self-attention mechanism to calculate asymmetric attention score matrix and applied it to sparse graph convolution.

Although the GAT-based approaches have made great progress in trajectory prediction, they require a huge amount of data to learn the differences between nodes in the graph and consume much more computing resources. The method proposed in our work is more hand-designed and has faster speed than that of the GAT-based method. Due to the thorough consideration, our method can better distinguish  different influences between various nodes.

\begin{figure*}[t]
	\begin{center}
		\includegraphics[width=1.0\linewidth]{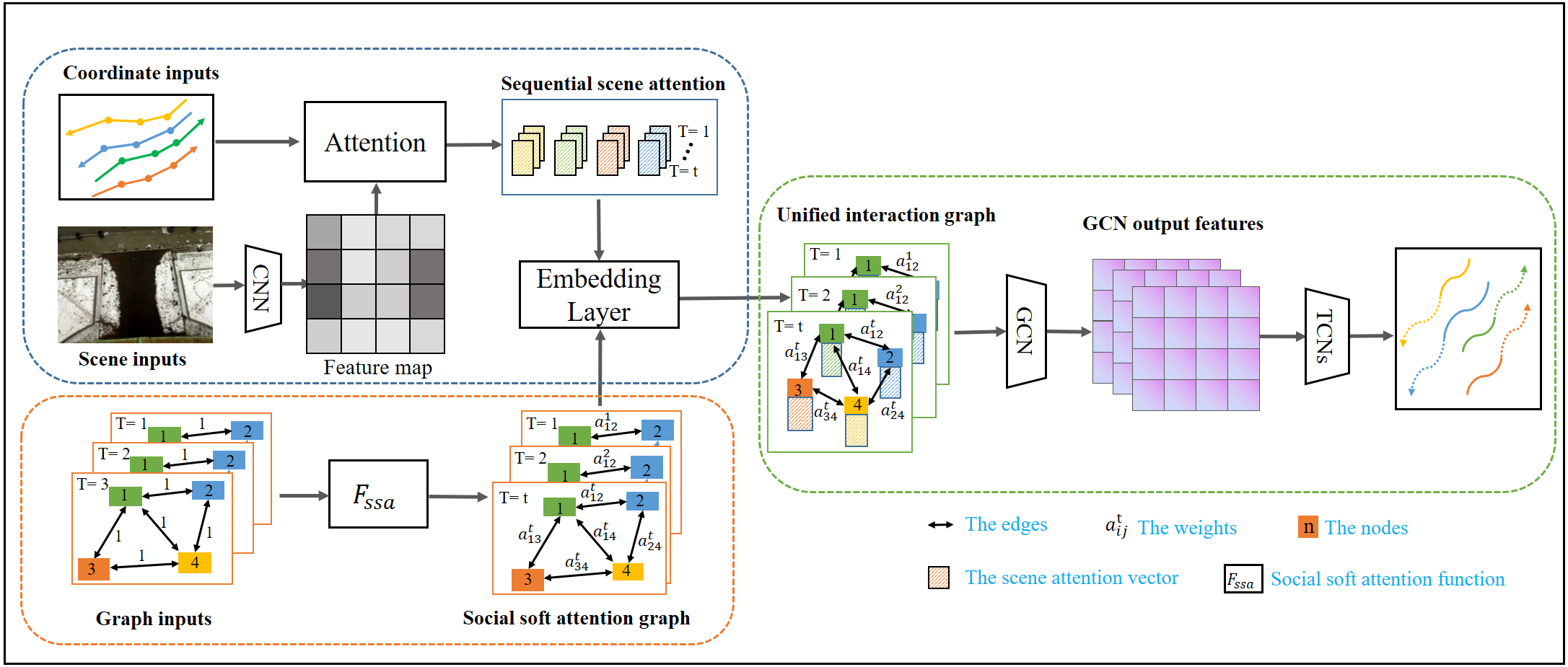}
	\end{center}
	\caption{The architecture of our proposed SSAGCN model. Our method merges the initial graph built using coordinates with the sequential scene attention, and then aggregates the social features and shares the physical scene influence under the action of the social soft attention function and GCN. Finally, the TCNs obtains the predicted trajectories. The arrows in both directions represent the edges between nodes, and the characters on the edges represent weights.}
	\label{fig2}
\end{figure*}

\section{SSAGCN}

In this section, we aim to develop a prediction model to derive a set of possible future trajectories for pedestrians. In Section~\ref{pro_def}, we explain the problem definition of pedestrian trajectory prediction. In Section~\ref{model_overview}  we give a brief overview of the proposed model. In the following Section~\ref{sagr}, we introduce the details of the design of social soft attention function. In Section~\ref{ssasm} we demonstrate the sequential scene attention sharing mechanism. Finally, the method of trajectory generation is introduced in Section~\ref{trj_gen}.

\subsection{Problem definition}
\label{pro_def}

Similar to~\cite{B3, A10}, we assume that there are $N$ pedestrians during one period of time $[1,{{T}_{pred}}]$. After pre-processing the trajectories of pedestrians in the video, the position of each pedestrian $i$ at each time step $t$ can be expressed as a pair of spatial coordinates $\left(x_{i}^{t}, y_{i}^{t}\right)$, where $t \in\{1,2,3, \ldots, T_{pred}\}$ , $i \in\{1,2,3, \ldots, N\}$. The scene image is represented by ${I}_{scene}$. We take the coordinate sequences and the scene information of ${I}_{scene}$ during the period of time interval $[1,{{T}_{obs}}]$ as the input, and predict the coordinate sequences in $[{{T}_{obs+1}},{{T}_{pred}}]$.
\subsection{Model overview}
\label{model_overview}

As illustrated in Figure~\ref{fig2}, firstly, we process the coordinates $\left(x_{i}^{t}, y_{i}^{t}\right)$ into graph representation, with each node representing an agent. Then we estimate the influence matrix among graph nodes using social soft attention function according to the relative position, speed and direction of agents. Meanwhile, scene attention $C_{i}^{t}$ is calculated according to the coordinates  $\left(x_{i}^{t}, y_{i}^{t}\right)$ of agent and the features of the scene image ${I}_{scene}$ at each past moment. The scene attention $C_{i}^{t}$ will be further embedded into graph node at each time step.

Through the implementation of GCN and social soft attention, scene information is shared among agents through social interaction. At last, we input the graph sequence processed by GCN into the TCNs~\cite{E15} to estimate the future trajectory of the agent. We will explain the details in the following sections.

\begin{figure}[t]
	\begin{center}
		\includegraphics[width=1.0\linewidth]{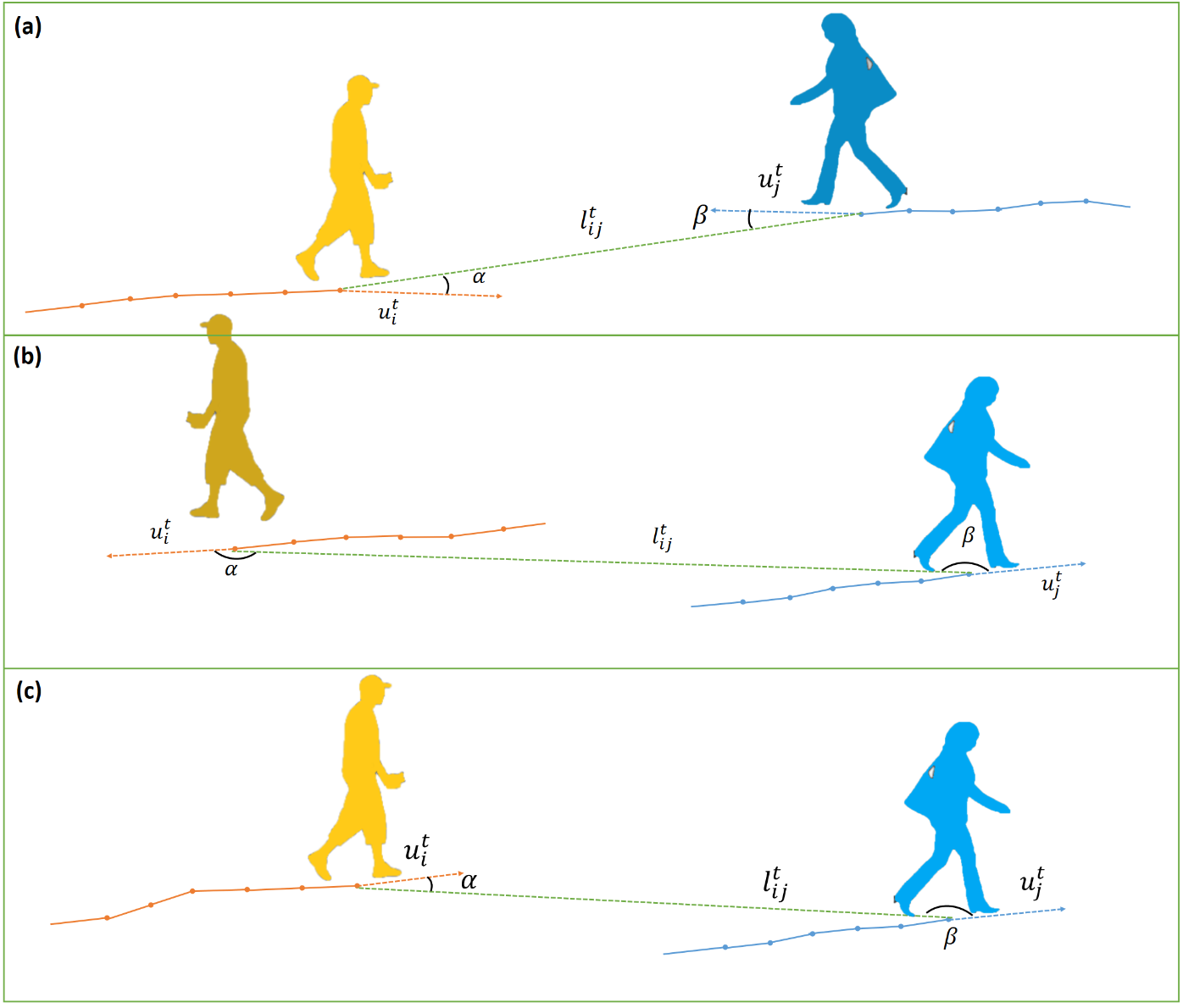}
	\end{center}
	\caption{Three types of pedestrian interaction. $u^{t}$ indicates the speed vector of the node $v^{t}$ at time step $t$, $\alpha$ and $\beta$ represent the angles of the line between the velocity vector and the agent. $l_{ij}^{t}$ represents the Euclidean distance between $v_{i}^{t}$ and $v_{j}^{t}$.} 
	\label{fig3}
\end{figure}

\subsection{Social attention graph representation}
\label{sagr}

\par \noindent \textbf{Node representation:} 
We construct a sequence of graphs to represent the pedestrian trajectories. At time step $t$, all pedestrians are connected to form a complete graph $G_{t}=\left(V_{t}, E_{t}\right)$. $V_{t}=\left\{v_{i}^{t} \mid \forall i \in\{1, \ldots, N\}\right\}$ is the set of vertices of $G_{t}$, representing all pedestrians at time step $t$. The initial value of $v_{i}^{t}$ is the observed coordinate position $\left(x_{i}^{t}, y_{i}^{t}\right)$. $E_{t}=\left\{e_{ij}^{t} \mid \forall i,j \in\{1, \ldots, N\}\right\}$ is the set of edges of $G_{t}$. In our work, the edges contained in $E_{t}$ are represented by an adjacency matrix $A_{t}$ calculated according to the social soft attention function $F_{ssa}\left(\cdot\right)$.


\par \noindent \textbf{Social soft attention function:}
According to the actual situation, the interactions among pedestrians can be divided into three categories as shown in Figure~\ref{fig3}: meeting from opposite directions, leaving in the opposite direction, and walking abreast. In order to cover these typical situations, we propose one new social soft attention function $F_{ssa}(\cdot)$ to calculate the element $a_{ij}^{t}$ in $A_{t}$ to represent the attention weight between pedestrians. The process is expressed as follows:
\begin{equation}
\begin{aligned}
a_{ij}^{t}&=F_{s s a}\left(u_{i}^{t}, u_{j}^{t}, \cos \alpha^{t}, \cos \beta^{t}, l_{i j}^{t}\right)\\
          &=\left\{\begin{array}{c}
\max \left(0, \frac{\left|u_{i}^{t}\right| \cos \alpha^{t}+\left|u_{j}^{t}\right| \cos \beta^{t}}{l_{i j}^{t}}\right), i \neq j \\
\theta, \quad \quad \quad \quad \quad \quad\quad\quad\quad \quad\quad\quad i=j
\end{array}\right.
\end{aligned}
\label{eq5}
\end{equation} 

\begin{equation}
\tilde{a}_{i j}^{t}=\operatorname{softmax}\left(a_{i j}^{t}\right)=\frac{e^{a_{i j}^{t}}}{\sum_{k=1}^{j} e^{a_{i k}^{t}}}
\label{eq6}
\end{equation}
 As shown in Figure~\ref{fig3}, $u_{i}^{t}$ indicates the speed vector of node $v_{i}^{t}$ at time step $t$, which is obtained by the $\left(x_{i}^{t}-x_{i}^{t-1}, y_{i}^{t}-y_{i}^{t-1}\right)$. Similarly, $u_{j}^{t}$ represents the speed vector of the node $v_{j}^{t}$ at time step $t$. $\alpha$ and $\beta$ represent the angle between the speed vectors $u_{i}^{t}$ of the node $v_{i}^{t}$ and $u_{j}^{t}$ of the node $v_{j}^{t}$, respectively. The $l_{ij}^{t}$ represents the Euclidean distance between node $v_{i}^{t}$ and node $v_{j}^{t}$. $\theta$ is a hyperparameter, which represents the self-attention of one agent. When we implement the social soft attention function, the value  of $\theta$ is set into the interval $[0.04,0.16]$ depending on the maximum value of elements in the matrix $A_{t}$. Equation \ref{eq6} is the normalization process to obtain the new adjacency matrix $\tilde{A}_{t}$. 
\par In Figure~\ref{fig3}(a), two people are facing each other. $\alpha$ and $\beta$ are both acute angles. The output of $F_{ssa}(\cdot)$ is larger than 0, which means these two people have an influence on each other. In Figure~\ref{fig3}(b), the two people  are leaving in the opposite directions. $\alpha$ and $\beta$ are both obtuse angles. The output of $F_{ssa}(\cdot)$ is limited to 0, which indicates the two people do not influence each other. In Figure~\ref{fig3}(c), two people are walking abreast, where $\alpha$ is an acute angle and $\beta$ is an obtuse angle. If the speed of the yellow pedestrian is much higher than that of the blue one, there will be a risk of collision in the future, then the output of $F_{ssa}(\cdot)$ is larger than 0. If the speed of yellow pedestrian is less than that of the blue one, there will be no collision risk, and the output of $F_{ssa}(\cdot)$ is limited to 0. We put the distance $l_{ij}^{t}$ in the denominator to reflect the magnitude of the risk of collision. When $\alpha$ and $\beta$ take different values, even 0° or 180°, our function still can distinguish different situations of pedestrian interaction, so that the features of social interaction can be better extracted by graph convolutional network.
\subsection{Sequential scene attention sharing mechanism}
\label{ssasm}
\noindent \textbf{The extended scene attention:} To fully leverage scene information, we utilize the pre-trained CNN network to extract the features of scene images. The method adopted in our work is similar to ~\cite{B9}. We use the pre-trained VGG19~\cite{B8} as the backbone network, and the extracted features are represented by ${V}_{p h}$. Since the images from the dataset  are captured by fixed cameras, we only need to calculate a ${V}_{p h}$ for each dataset. In previous work~\cite{B8,B15}, they usually chose to calculate the scene attention $C_{i}^{{T}_{obs}}$ when $t= {T}_{obs}$. In order to extend the influence of the scene, we calculate the scene attention $C_{i}^{t}$ at each moment $t \in\{1,2,3, \ldots, {T}_{obs}\}$ according to the feature ${V}_{p h}$ and the position $X_{i}^{t}$ of agent $i$.

\begin{equation}
	V_{p h}=V G G 19\left(I_{scene} ; W_{v g g 19}\right)
\end{equation}

\begin{equation}
	C_{i}^{t}=\operatorname{SceneAtt}\left(V_{p h}, X_{i}^{t} ; W_{a t t}\right)
\end{equation}

\noindent here, ${I}_{scene}$ represent the images from datasets, ${W}_{vgg19}$ is the weight of pre-training network, $X_{i}^{t}$ is the position $\left(x_{i}^{t}, y_{i}^{t}\right)$ of the agent $i$ at time step $t$, and ${W}_{att}$ contains the parameters of the scene attention.
Then we embed $C_{i}^{t}$ with coordinate information to form new graph node feature $v_{i}^{t}$. 
\begin{equation}
	v_{i}^{t}=\phi\left(X_{i}^{t},C_{i}^{t};W_{e}\right)
\label{eqs}
\end{equation}
\noindent where $\phi(\cdot)$ is the embedding layer and $W_{e}$ is the weight of embedding layer.
\par \noindent \textbf{Graph neural network:} With the above graph representation, we perform graph convolution operation at each time step $t$. The convolution operation of graph at  single moment $t$ is defined in GCN~\cite{D1} as following:

\begin{equation}
\tilde{\mathcal{A}}_{t}=\mathcal{A}_{t}+E
\label{eq1}
\end{equation}
\begin{equation}
\tilde{D}_{t}=D_{t}+E
\label{eq2}
\end{equation}

\begin{equation}
{V_{t}}^{\prime}=\sigma\left(\tilde{D}_{t}^{-\frac{1}{2}} \tilde{\mathcal{A}}_{t} \tilde{D}_{t}^{-\frac{1}{2}} V_{t} W\right)=\sigma\left(\tilde{L}_{t} V_{t} W\right)
\label{eq3}
\end{equation}

    \noindent where $\mathcal{A}_{t}$ is the adjacency matrix of the graph at time step $t$, $D_{t}$ is the degree matrix of the graph, and $E$ is the identity matrix; $V_{t}$ is a matrix composed of $v_{i}^{t}$  as row vectors at the same time step. $\tilde{L}_{t}$ is a graph displacement operator used to aggregate adjacent nodes, which is the result of normalization of $\mathcal{A}_{t}$.

\par For a single node, the graph convolution operation is:
\begin{equation}
{v_{i}^{t}}^{\prime}=\sigma\left(\sum_{v_{j}^{t} \in N\left(v_{i}^{t}\right)} \widetilde{L}_{t}[i, j]\left(W v_{j}^{t}\right)\right)
\label{eq4}
\end{equation}
\noindent where $N\left(v_{i}^{t}\right)$ is the set of first-order neighbors of node $v_{i}^{t}$, $\tilde{L}_{t}[i, j]$  is the element in the $i$ row and $j$ column of matrix $\tilde{L}_{t}$. During the process of aggregating neighbor features of $v_{i}^{t}$, $\tilde{L}_{t}[i, j]$ represents the weight of $v_{j}^{t}$.
\par \noindent \textbf{Scene attention sharing} Since the node features $v_{i}^{t}$ involved in the graph convolution operation is embedded with $C_{i}^{t}$ in Equation \ref{eqs}, the process of scene attention sharing of agents can be expressed as:
\begin{equation}
{C_{i}^{t}}^{'}=\sigma\left(\sum_{C_{j}^{t} \in N\left(C_{i}^{t}\right)} \widetilde{L}_{t}[i, j]\left(W \phi\left(X_{i}^{t},C_{j}^{t};W_{e}\right)\right)\right)
\label{eq10}
\end{equation}

\begin{figure}[t]
	\begin{center}
		\includegraphics[width=0.9\linewidth]{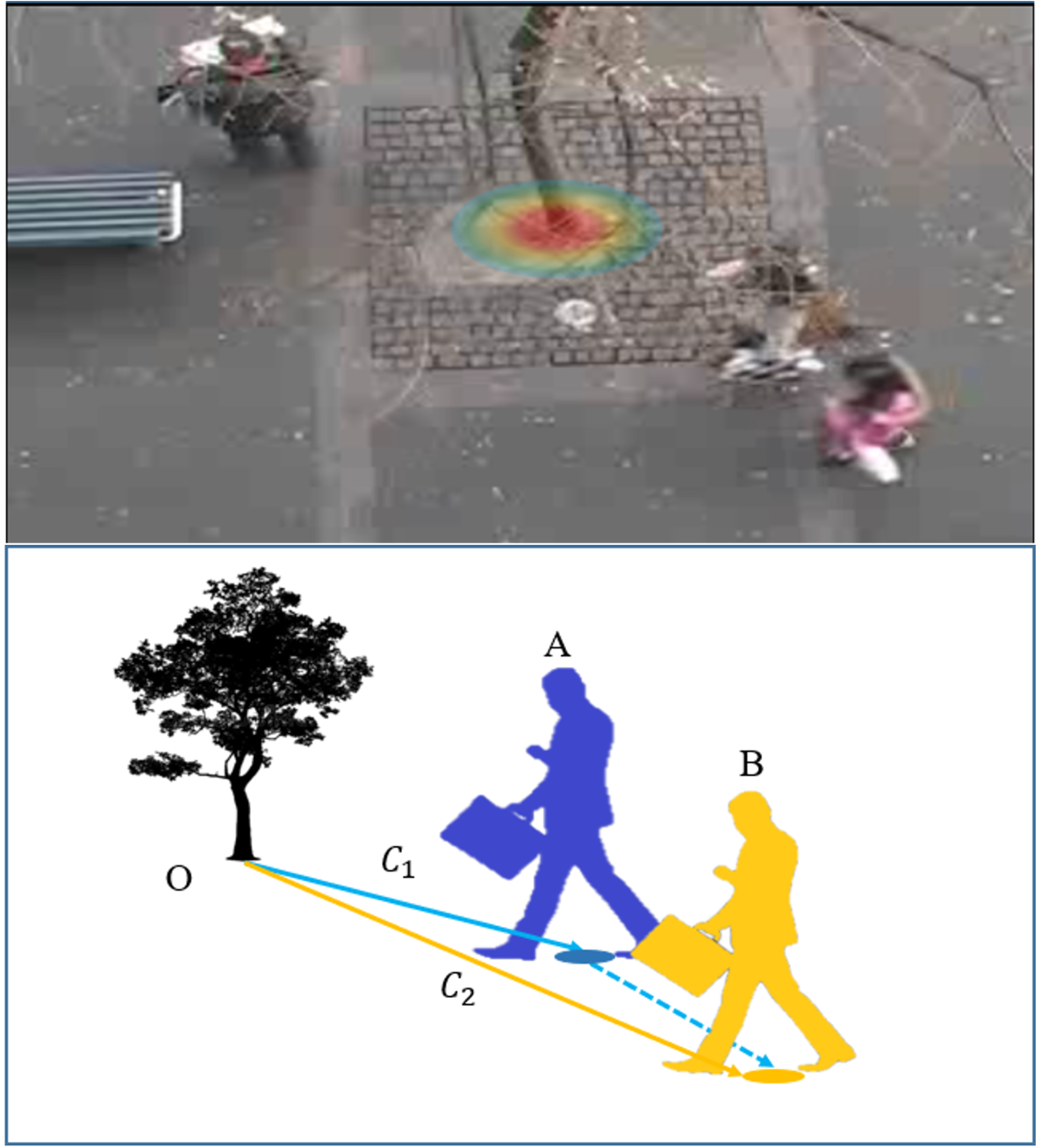}
	\end{center}
	\caption{One demonstration of the attention-sharing mechanism in certain scenario. The top part of this figure is the real scene, and the bottom part is the corresponding simulated one. $C_{1}$ and $C_{2}$ represent the effect of obstacles O on A and B. The dotted blue line represents part of the effect of the obstacle on A is transmitted to B through the social interaction between A and B.}
	\label{figs}
\end{figure}

The impact of scene on agents during this process is shared through social interactions. In Figure~\ref{figs}, the direct effect of the barrier on agent A is $C_1$, and that of the barrier on agent B is $C_2$. In the sharing process, the effect of obstacles on A is transmitted to B through the social relationship between A and B. The impact of obstacles on Agent B is aggregated into $C_{2}'=C_{1}+ WW_{e}C_{2}$.

In order to better distinguish the interaction between different pedestrians, after adding the social soft attention function, Equation \ref{eq4} is rewritten as:
\begin{equation}
V_{t}^{\prime}=\sigma\left(\tilde{A}_{t} V_{t} W\right)
\end{equation}
where the $\tilde{A}_{t}$ is given by Equation \ref{eq6}.

\subsection{Trajectory generation}
\label{trj_gen}
After dealing with social interaction and scene interaction, we choose TCNs to model the temporal dependence of social graph sequences $V_{t}^{\prime} =\left\{{v_{i}^{t}}^{\prime} \mid \forall i \in\{1, \ldots, N\}\right\}$. Just like Social-STGCNN, we treat the time dimension as feature channels input ${v_{i}^{t}}^{\prime}$ into TCNs. After the convolution operation of TCNs, we obtain the parameters of the two-dimensional Gaussian distribution about the sequence of coordinates during $[{{T}_{obs+1}},{{T}_{pred}}]$. 

\begin{equation}
\left[\mu_{x}^{\tau}, \mu_{y}^{\tau}, \sigma_{x}^{\tau}, \sigma_{y}^{\tau}, \rho^{\tau}\right]=TCNs\left(V_{t}^{\prime}, W_{c}\right)
\end{equation}
\noindent where $\mu_{x}^{\tau},\mu_{y}^{\tau}$ are the mean of the coordinates and $\sigma_{x}^{\tau},\sigma_{y}^{\tau}$ are variance. $\rho^{\tau}$ is the correlation coefficient between $x$ and $y$. $W_{c}$ is the weight of TCNs. $\tau$ are time steps in $[{{T}_{obs+1}},{{T}_{pred}}]$. Then we construct a two-dimensional Gaussian distribution $\mathcal{N}$ by $\left[\mu_{x}^{\tau}, \mu_{y}^{\tau}, \sigma_{x}^{\tau}, \sigma_{y}^{\tau}, \rho^{\tau}\right]$, and the future coordinates can be obtained by sampling the distribution. \begin{equation}
(\hat{{x}_{i}^{\tau}}, \hat{{y}_{i}^{\tau}})\sim \mathcal{N}\left(\mu_{x}^{\tau}, \mu_{y}^{\tau}, \sigma_{x}^{\tau}, \sigma_{y}^{\tau}, \rho^{\tau}\right)
\end{equation}

\par \noindent \textbf{Loss Function}
We use the ground truth values ${x}_{i}^{\tau}, {y}_{i}^{\tau}$ and the predicted parameters of Gaussian distribution to calculate the negative log-likelihood loss to guide the model training.
\begin{equation}
\mathcal{L}= -\sum_{t=T_{o b s}+1}^{T_{p r e d}} \log \left(\mathbb{P}\left(x_{i}^{\tau}, y_{i}^{\tau} \mid \sigma_{t}^{i}, \mu_{t}^{i}, \rho_{t}^{i}\right)\right)
\end{equation}
\section{Experiment and evaluation}

We perform the evaluation experiments on widely used benchmark datasets and compare the results with other state-of-the-art methods. In detail, we use ETH~\cite{A19} and UCY~\cite{A20} datasets, which contain the following scenarios, ETH, HOTEL, UNIV, ZARA1 and ZARA2. The data attributes consist of frame number, pedestrian number, and 2D position of trajectory coordinates. These trajectories are spaced 0.4 seconds apart from each other. Similar to Social-LSTM~\cite{B3}, we also input the trajectory of 8 time steps (3.2 seconds) and predict the next 12 time steps (4.8 seconds). To verify the proposed method is able to deal with various scenarios, we also conduct experiments on Stanford Drone dataset(SDD)~\cite{E12}, which contains a large number of different scenes. The tracking coordinates in this dataset are measured in pixels. We use the same standard data segmentation settings as~\cite{E4}.

\subsection{Experiment}
\begin{table*}[htbp]
	\begin{center}
	\caption{Quantitative comparison with baseline methods on ETH and UCY datasets. All these methods are used to predict the trajectory of the future 12 frames based on the previous 8 frames. The evaluation metrics used in this table are ADE and FDE. $K$ represents the number of predicted trajectories. The data in this table are from the results reported in their work. SSAGCN-w/o-sen, SSAGCN-w/o-seq and SSAGCN-w/o-ssa are our models with different experimental settings.}
	\label{tab1}
	\resizebox{\linewidth}{!}{
	\begin{tabular}{c|c|c|c|c|c|c}
		\toprule
        
        \multicolumn{7}{c}{(a) ADE/FDE $K=1$ } \\
		\midrule
		&ETH&	HOTEL&	UNIV&	ZARA1&	ZARA2&	AVG\\
		\midrule
		S-LSTM~\cite{B3}&	1.09 / 2.35&	0.79 / 1.76&	0.67 / 1.40&	0.47 / 1.00&	0.56 / 1.17&	0.72 / 1.54\\
		SR-LSTM~\cite{B6}&	0.63 / 1.25&	0.37 / 0.74&	0.51 / 1.10&	0.41 / 0.90&	0.32 / 0.70&	0.45 / 0.94\\
		STGAT~\cite{B15}&	0.75 / 1.55&	0.43 / 0.88 &	0.31 / 0.66&	0.25 / 0.53&	0.21 / 0.44&	0.39 / 0.81\\
		GAT~\cite{A16}&	0.68 / 1.29&	0.68 / 1.40&	0.57 / 1.29&	0.29 / 0.60&	0.37 / 0.75&	0.52 / 1.07\\
		Social-BiGAT~\cite{A16}&	-&	-&	-&	-&-&	0.61 / 1.33\\
		RSBG~\cite{B21}&	0.80 / 1.53&	0.33 / 0.64 &	0.59 / 1.25&	0.40 / 0.86&	0.30 / 0.65&	0.48 / 0.99\\
		Star~\cite{E8}&	0.56 / 1.11&	0.26 / 0.50 &	0.40 / 0.89&	0.31 / 0.71&	0.52 / 1.13&	0.41 / 0.87\\
		SCAN~\cite{E9}&	0.57 / 0.78&	0.43 / 0.85 &	0.61 / 1.28&	0.39 / 0.84&	0.34 / 0.74&	0.46 / 0.89\\
		Trajectron++~\cite{E++}&	0.71 / 1.68&	\textbf{0.22} / 0.46 &	0.41 / 1.07&	0.30 / 0.77&	0.23 / 0.59&	0.37 / 0.91\\
		\midrule
		SSAGCN-w/o-sen &	0.50 / 0.87&	0.31 / 0.49&	0.35  / 0.72&	0.27 / 0.45&	0.24 / 0.36&	0.33 / 0.58\\
		SSAGCN-w/o-seq &	0.45 / 0.76&	0.29 / 0.52&	0.35  / 0.68&	0.25 / 0.43&	0.26 / 0.40&	0.32 / 0.51\\
		SSAGCN-w/o-ssa &	0.38 / 0.63&	0.25 / 0.48&	0.26 / 0.53&	0.22 / 0.42&	0.23 / 0.36&	0.27 / 0.49\\
        SSAGCN &	\textbf{0.30} / \textbf{0.59}&	\textbf{0.22} / \textbf{0.42}&	\textbf{0.25} / \textbf{0.47}&	\textbf{0.20} / \textbf{0.39}&	\textbf{0.14 / 	0.28}& \textbf{0.22 / 0.43}\\
        \midrule
		\multicolumn{7}{c}{(b) ADE/FDE $K=20$ } \\
		\midrule
		SGAN~\cite{A10}	&0.87 / 1.62&	0.67 / 1.37&	0.76 / 1.52&	0.35 / 0.68&	0.42 / 0.84&	0.61 / 1.21\\
		Sophie~\cite{B8}&	0.70 / 1.43&	0.76 / 1.67&	0.54 / 1.24&	0.30 / 0.63&	0.38 / 0.78&	0.54 / 1.15\\
		STGAT~\cite{E9}&	0.56 / 1.10&	0.27 / 0.50 &	0.32 / 0.66&	0.21 / 0.42&	0.20 / 0.40&	0.31 / 0.62\\
		Social-BiGAT~\cite{A16}& 0.69 / 1.29 & 0.49 / 1.01 &0.55 / 1.32	& 0.30 / 0.62 & 0.36 / 0.75 &	0.48 / 1.00\\
		NMMP~\cite{E4}&	0.62 / 1.08&	0.33 / 0.63 &	0.52 / 1.11&	0.32 / 0.66&	0.29 / 0.61&	0.41 / 0.82\\
		Social-STGCNN~\cite{C1}&	0.64 / 1.11&	0.49 / 0.85&	0.44 / 0.79&	0.34 / 0.53&	0.30 / 0.48&	0.44 / 0.75\\
		CARPe~\cite{E10}&	0.80 / 1.48&	0.52 / 1.00 &	0.61 / 1.23&	0.42 / 0.84&	0.34 / 0.74&	0.46 / 0.89\\
		PECnet~\cite{E6}& 0.54 / 0.87 & 0.18 / 0.24 & 0.35 / 0.60& 0.22 / 0.39 & 0.17 / 0.30 & 0.29 / 0.48\\
		SCAN~\cite{E9}&	0.84 / 1.58&	0.44 / 0.90 &	0.63 / 1.33&	0.31 / 0.85&	0.37 / 0.76&	0.51 / 1.08\\
		Trajectron++~\cite{E++}&	0.43 / 0.86&	0.12 / 0.19 &	0.22 / 0.43&	0.17 / 0.32&	0.12 / 0.25&	0.21 / 0.41\\
		GTPPO~\cite{E17}&	0.63 / 0.98&	0.19 / 0.30 &	0.35 / 0.60&	0.20 / 0.32&	0.18 / 0.31&	0.31 / 0.50\\
		SGCN~\cite{E16}&	0.52 / 1.03&	0.32 / 0.55 &	0.37 / 0.70&	0.29 / 0.53&	0.25 / 0.45&	0.37 / 0.65\\
		Introvert~\cite{E18}&	0.42 / 0.70&	0.11 / 0.17 &	0.20 / 0.32&	0.16 / 0.27&	0.16 / 0.25&	0.21 / 0.34\\
		 LB-EBM~\cite{E19}&	0.30 / 0.52&	0.13 / 0.20 &	0.27 / 0.52&	0.20 / 0.37&	0.15 / 0.29&	0.21 / 0.38\\
		Y-Net~\cite{E21}&	0.28 / \textbf{0.33}&	\textbf{0.10 / 0.14} &	0.24 / 0.41&	0.17 / 0.27&	0.13 / 0.22&	0.18 / 0.27\\
        \midrule
		SSAGCN-w/o-sen &	0.27 / 0.46&	0.20 / 0.27&	0.26  / 0.43&	0.17 / 0.34&	0.16 / 0.28&	0.21 / 0.36\\
		SSAGCN-w/o-seq &	0.26 / 0.41&	0.18 / 0.27&	0.24  / 0.41&	0.18 / 0.33&	0.15 / 0.26&	0.20 / 0.34\\
		SSAGCN-w/o-ssa &	0.23 / 0.41&	0.17 / 0.26&	0.19 / 0.36&	0.17 / 0.30 & 0.12 / 0.21&	0.18 / 0.31\\
        SSAGCN &	\textbf{0.21} / 0.38&	0.11 / 0.19&	\textbf{0.14 / 0.25}&	\textbf{0.12 / 0.22}&	\textbf{0.09 / 	0.15}&	\textbf{0.13 / 0.24}\\
		\bottomrule
	\end{tabular}
	}
    \end{center}
	\centering
	
\end{table*}

\begin{table*}[htbp]
\begin{center}
\caption{The evaluation results of SSAGCN on SDD taking ADE and FDE as evaluation criteria, and pixel as scale. The data in this table are from the results reported in their work.}
\label{tab3}
    \resizebox{\linewidth}{!}{
	\begin{tabular}{c|c|c|c|c|c|c|c|c}
	Metric&SGAN~\cite{A10}&Sophie~\cite{B8}&PMP-NMMP~\cite{E4}&GTPPO~\cite{E17}&PECnet~\cite{E6}&LB-EBM~\cite{E19}&Y-Net~\cite{E21}&SSAGCN	\\
			\midrule
ADE&	27.25&16.27&	14.67&10.13&9.96&8.87&	\textbf{7.85}&10.36	\\
FDE&    41.44&	29.38&	26.72&	15.35&	15.88&	15.61&11.85&	\textbf{11.80} \\
		\bottomrule

	\end{tabular}
	}
    \end{center}
	\centering
	
\end{table*}

\begin{table}[htbp]
\begin{center}
\caption{The average percentage of human colliding for each scene in ETH and UCY datasets. The human collision is defined in~\cite{B8}.}
\label{tab4}
	\begin{tabular}{c|c|c|c|c|c|c}    
		\toprule            
	&GT&	Liner&	SGAN&	Sophie& SCAN &	Ours\\
			\midrule 
ETH&	0.000&	3.137&	2.509&	1.757& \textbf{0.793} &	1.145\\
HOTEL&	0.092&	1.568&	1.752&	1.936& 1.126&	\textbf{0.573}\\
UNIV&	0.124&	1.242&	0.559&	0.621& 0.481&	\textbf{0.277}\\
ZARA1&	0.000&	3.776&	1.749&	1.027& 0.852&	\textbf{0.634}\\
ZARA2&	0.732&	3.631	&2.020&	1.464&  3.109&	\textbf{1.428}\\
		\midrule 
Avg	&0.189&	2.670&	1.717&	1.361&  1.272&	\textbf{0.811}\\
		\bottomrule         
     
	\end{tabular}
    \end{center}
	\centering
	
\end{table}

\par\noindent \textbf{Evaluation criteria.} Following the strategy adopted by other baseline methods, we use the leave-one-out method to conduct the evaluation experiments, train on four datasets and test on the remaining one. Average Displacement Error (ADE) and Final Displacement Error (FDE) are used as the standard metrics, which are defined as:
\begin{equation}
ADE=\frac{\sum\nolimits_{n\in N}{\sum\nolimits_{t\in {{T}_{pred}}}{||\hat{Y}_{n}^{t}-Y_{n}^{t}|{{|}_{2}}}}}{N*{{T}_{pred}}}
\label{eq14}
\end{equation}
\begin{equation}
FDE=\frac{\sum\nolimits_{n\in N}{||\hat{Y}_{n}^{t={{T}_{pred}}}-Y_{n}^{t={{T}_{pred}}}|{{|}_{2}}}}{N}
\label{eq15}
\end{equation}

\par \noindent\textbf{Implementation details.} The structure of GCN used in SSAGCN is similar to~\cite{C1}. We use coordinate information in data processing to calculate the weighted adjacency matrix through the social soft attention function. When the weighted adjacency matrix is normalized, $R$ is set to 0.10. The dimension of scene attention is 8, and the coordinate dimension is 2. Therefore, the input dimension of the embedding layer is 10. We set the output dimension of the embedding layer as 5, corresponding to the number of parameters in the Gaussian distribution. The input and output dimensions of TCNs correspond to the time dimensions of the input sequence and the prediction sequence respectively. To avoid over smoothing, we only use one layer of GCN and six layers of TCNs. Finally, we use the SGD optimizer and learning rate with 0.001 to train the SSAGCN model for 200 epochs on Tesla V100 GPU.

\subsection{Quantitative analysis}
\par We choose the state-of-the-art methods as the baselines.
\par\noindent\textbf{S-LSTM~\cite{B3}:} A LSTM network based on prediction method using an ingenious pooling mechanism.
\par\noindent\textbf{SGAN~\cite{A10}:} One method introducing GAN into pedestrian trajectory prediction and using global pooling for interaction.
\par\noindent\textbf{SR-LSTM~\cite{B6}:} A prediction method using states refinement for LSTM network.
\par\noindent\textbf{Sophie~\cite{B8}:} GAN-based prediction method considering both scene factors and social factors using attention mechanism.
\par\noindent\textbf{STGAT~\cite{B15}:} A Spatial-Temporal Graph Attention network based on a sequence-to-sequence architecture to predict future trajectories of pedestrians.
\par\noindent\textbf{Social-BiGAT~\cite{A16}:} One method adopting a cyclic confrontation structure and introducing one graph attention network to calculate the impact between pedestrians.
\par\noindent\textbf{Social-STGCNN~\cite{C1}:} One method modeling pedestrian trajectories into graphs and using STGCN to deal with social interactions.
\par\noindent\textbf{RSBG~\cite{B21}:} A recursive social behavior graph combined with GCN is proposed to model social interaction.
\par\noindent\textbf{NMMP~\cite{E4}:} A neural motion message passing is proposed for interactive modeling, which can predict future trajectories in a variety of scenarios.
\par\noindent\textbf{Star~\cite{E8}:} A novel spatial graph transformer is introduced to capture the interaction between pedestrians.
\par\noindent\textbf{PECNet~\cite{E6}:} A two-stage prediction framework used to predict the end point of the trajectory, and then a reasonable path planning.
\par\noindent\textbf{SCAN~\cite{E9}:}  A Spatial Context Attentive Network that can jointly predict socially-acceptable multiple future trajectories for all pedestrians in a scene. 
\par\noindent\textbf{CARPe~\cite{E10}:} A convolutional approach for real-time pedestrian path prediction, which utilizes a variation of Graph Isomorphism Networks in combination with an agile convolutional neural network design.
\par\noindent\textbf{SGCN~\cite{E16}:} A pedestrian path prediction method using sparse graph convolution and self-attention mechanism to calculate asymmetric attention score matrix.
\par\noindent\textbf{GTPPO~\cite{E17}:} A graph based Pseudo-Oracle trajectory Predictor (GTPPO), which encodes pedestrian movement patterns using short and long memory units and introduces temporal attention to highlight specific temporal steps. 
\par\noindent\textbf{Trajectron++~\cite{E++}:} An approach designed to be tightly integrated with robotic planning and control frameworks, which can produce predictions that are optionally conditioned on ego-agent motion plans.
\par\noindent\textbf{Introvert~\cite{E18}:} A pedestrian trajectory prediction method using 3D visual attention mechanism to capture dynamic scene context.
\par\noindent\textbf{LB-EBM~\cite{E19}:} A latent belief energy-based model (LB-EBM) for diverse human trajectory forecast.
\par\noindent\textbf{Y-Net~\cite{E21}:} A scene compliant trajectory forecasting model that exploits the proposed epistemic and aleatoric structure for diverse trajectory predictions across long prediction horizons.
\par\noindent\par We compare the experimental results with above baseline methods, and their ADE and FDE are shown in Table~\ref{tab1}. Compared with these methods, our method achieves the best performance at $K=1$ and $K=20$. $K=1$ means that the model generates only one trajectory, while $K=20$ means that the model predicts 20 trajectories or samples 20 times. It can be seen from Table~\ref{tab1} that the Trajectron++ has been at the forefront in the past due to its advanced predictive framework and unique modeling approach, and has a decrease of nearly $45\%$ on ADE and FDE compared to the methods during the same period. Recent methods such as SGCN, Introvert and LB-EBM provide new research ideas for trajectory prediction. SGCN introduces sparse graph convolution network and combines with self-attention mechanism, which reduced ADE and FDE by $15\%$ compared with other GCN based methods. Introvert converts trajectory prediction to 3D domain, focusing on dynamic scene context, with a $17\%$ FDE reduction compared to Trajectron++. Our model uses a customized social soft attention function, which covers various pedestrian social interaction factors, enabling our model to better learn the social interaction between pedestrians. In addition, our model has the additional input of the scene image, and extends the role of the scene in time and space. As a result of the above novel contributions our model achieved state-of-the-art results, ADE reduction of $17\%$ and FDE reduction of $11\%$ compared to the previous baselines.


\par Table~\ref{tab3} shows the performance of our approach compared to others on the SDD dataset. It can be seen that the performance of SSAGCN on SDD is also competitive compared with other baseline methods, and the FDE is at its lowest level. In fact, the obstacle and road information is abundant in SDD dataset, and the input of scene information can provide great help for trajectory prediction. The Y-Net model contains the input of the scene image, and its results are $20\%$ less in ADE and $25\%$ less in FDE than previous methods. Our SSAGCN also takes as the input scene image and enhances the role of physical scene in spatial and temporal space, which makes our prediction result is 0.05 lower than that of Y-NET on FDE.

\par In addition, we use the percentage of near-collisions (whether the distance between two pedestrians is less than 0.1m) used by SoPhie~\cite{B8} to further evaluate our results, which are shown in Table~\ref{tab4}. In the aspect of this new evaluation methodology, our method is completely beyond other methods, which indicates it can generate better socially and physically acceptable trajectories for each pedestrian.

\subsection{Ablation study}
\par We set up two groups of comparative experiments to prove the effectiveness of the new social soft attention function. The SSAGCN-w/o-ssa in Table~\ref{tab1} shows the results obtained by directly using the adjacent matrix $A_{i}^{t}$ to aggregate node feature without our social soft attention function. The results show that the role of social soft attention function is positive. In addition, we further use social soft attention function instead of the kernel function of Social-STGCNN. The results are shown in Table~\ref{tab5}. Most of the results are better than the original Social-STGCNN on ADE and FDE.
\begin{table}[htbp]
	\begin{center}
	\caption{The comparison results obtained by Social-STGCNN with different kernel functions on ETH and UCY datasets. The first column is the original version of Social-STGCNN (source code is downloaded from~\url{https://github.com/abduallahmohamed/Social-STGCNN}.). The second column is the Social-STGCNN with our social soft attention function.}
\label{tab5}
    
	\begin{tabular}{c|c|c}
    &SSTGCNN w/o&	SSTGCNN w/\\
		\midrule
ETH  &0.81/1.58 &0.81/1.24 \\
HOTEL& 0.47/0.76 &0.37/0.59\\
UNIV &0.53/0.95&0.50/0.89\\
ZARA1 &0.38/0.60 &0.38/0.62\\
ZARA2 &0.32/0.53 &0.31/0.51\\
AVG&0.50/0.88 &0.47/0.77\\
		\bottomrule
	\end{tabular}
	
   \end{center}
	\centering
	
\end{table}

\begin{figure}[htb]
	\begin{center}
		\includegraphics[width=0.8\linewidth]{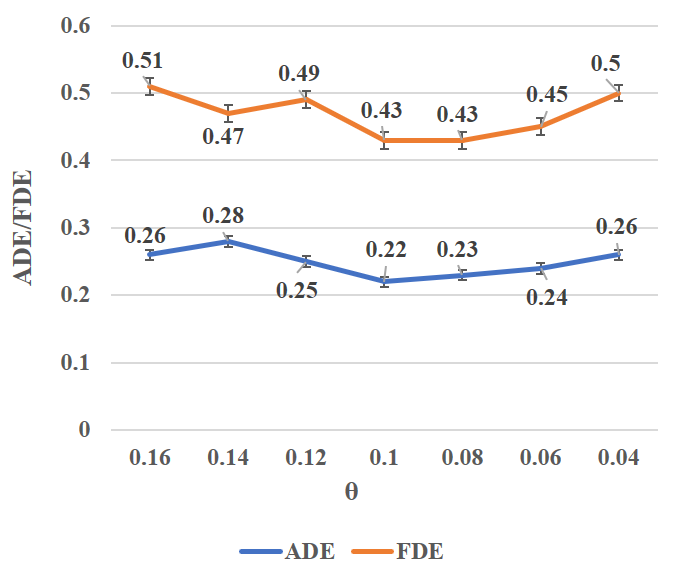}
	\end{center}
	\caption{The ablation results on $\theta$. Seven different values are used to show the model performance.}
\label{fig5}
\end{figure}

\begin{table*}[ht]
	\small
	\begin{center}
		\caption{Ablation results for the function $F_{ssa}$. The Speed, Direction, and Distance correspond to the variables in Equation~\ref{eq5}. The $\surd$ indicates the variable is considered in the $F_{ssa}$ during the experiment, and $\times$ indicates it is not involved.}
    \label{Fssa}

	\begin{tabular}{c|c|c|c|c|c}
Row& Speed & Direction&Distance& Average ADE& Average FDE\\
		\midrule
1& $\times$ &$\times$&$\times$& 0.33& 0.58\\
2& $\surd$ &$\times$&$\times$& 0.31& 0.57\\
3& $\times$ &$\surd$&$\times$& 0.30& 0.56\\
4&  $\times$ &$\times$&$\surd$& 0.31& 0.54\\
5&  $\surd$ &$\surd$&$\times$& 0.28& 0.50\\
6&   $\surd$ &$\times$&$\surd$& 0.27& 0.51\\
7& $\times$ &$\surd$&$\surd$& 0.25& 0.47\\
8& $\surd$ &$\surd$&$\surd$& \textbf{0.22}& \textbf{0.43}\\
		\bottomrule
	\end{tabular}
	
   \end{center}
	\centering
\end{table*}

\begin{table}[htbp]
	\begin{center}
		\caption {Comparative experiments on the efficiency of graph structure-based methods. The first row shows the number of parameters contained in each model. The remaining lines are the time taken by each model to predict the trajectory on different datasets.}
    \label{tab7}
    \resizebox{\linewidth}{!}{
	\begin{tabular}{c|c|c|c|c|c}
    & STGAT& SSAGAT& SGCN& SSTGCNN& SSAGCN\\
		\midrule
 Parm  & 44,630 &  12,575& 25,369& \textbf{7,563}& 7,578 \\
ETH  & \textbf{4.60s} & 9.53s & 5.15s & 4.63s  & 5.00s\\
HOTEL & 9.93s & 15.34s & 6.23s & \textbf{4.54s} & 4.58s\\
UNIV & 29.65s & 40.44s & 22.25s & 19.54s & \textbf{17.10s}\\
ZARA1 & 18.13s & 24.25s & 9.70s & 6.75s & \textbf{6.63s}\\
ZARA2 & 26.88s  & 34.54s & 14.39s & 9.74s & \textbf{9.62s}\\
AVG & 17.84s & 24.82s & 11.54s &  9.04s & \textbf{8.59s}\\
		\bottomrule
	\end{tabular}
	}
   \end{center}
	\centering

\end{table}

\par In addition, when we implement the social soft attention function, we use a hyperparameter $\theta$, to specify the values of the diagonal elements in the matrix $A_{i}^{t}$. The value of the diagonal element refers to the impact of pedestrians on themselves. When we implement SSAGCN, we take the $\theta$ value in the interval $[0.04,0.16]$, the mid-value of which is the maximum value of the off-diagonal element calculated by the social soft attention function. In Figure \ref{fig5}, we take seven different values of $\theta$ from the above interval for experiment. The performance of the model increases as $\theta$ decreased from 0.16 to 0.1. The model starts to look bad as we go down further. As $\theta$ continues to decrease from 0.1, the performance of the model degrades.

\par In order to prove the rationality of social soft attention function, we used different combinations of speed, direction and distance to conduct ablation experiments. The experimental results are shown in Table~\ref{Fssa}. It can be seen that better results can not be achieved by simply considering more factors. The influence between nodes is still dominated by the distance between them. However, it is not reasonable to distinguish the influence of different nodes only by distance. Our function achieves better results by combining more influences in a suitable form.

\par In order to further verify that our method is more effective in exploring scene information, we implement the ablation experiment of sequential scene attention sharing mechanism (shown in Table~\ref{tab1}). SSAGCN-w/o-sen in Table~\ref{tab1} shows the evaluation results without the sequential scene attention sharing and the SSAGCN is the result of our complete model. It can be seen that due to the efficient scene sharing, the average ADE and FDE are increased by $16$\% and $11$\% respectively. In addition, in order to further demonstrate the advantage of sequential scene attention, we also designed an experimental setup called STGCNN-w/o-seq that used only the last frame of the historical trajectory to calculate the impact of the scene and compare it to the full version. The results show that it is more advantageous to consider the impact from sequential attention sharing not just one single frame.

\par The structure of GCN and social soft attention function in our method is based on prior knowledge. Compared with learning methods such as GAT and SGCN, it has fewer parameters and is faster. We conduct a set of experiments to replace the GCN and the social soft attention function in our method with GAT. Since SGCN and Social-STGCNN (SSTGCNN) do not contain modules about the scene, the result of our model shown in Table \ref{tab7} is also the version without the scene information. It can be seen that the methods of modeling pedestrian interaction based on prior knowledge, SSAGCN and STGCNN, have fewer parameters and faster inference speed than that using attention mechanism in the previous columns. It is to be mentioned our model and SSTGCNN have similar number of parameters, but the prediction speed of our model is faster than other methods in this table.

\subsection{Qualitative analysis}

\begin{figure*}[htp]
	\begin{center}
		\includegraphics[width=1.0 \linewidth]{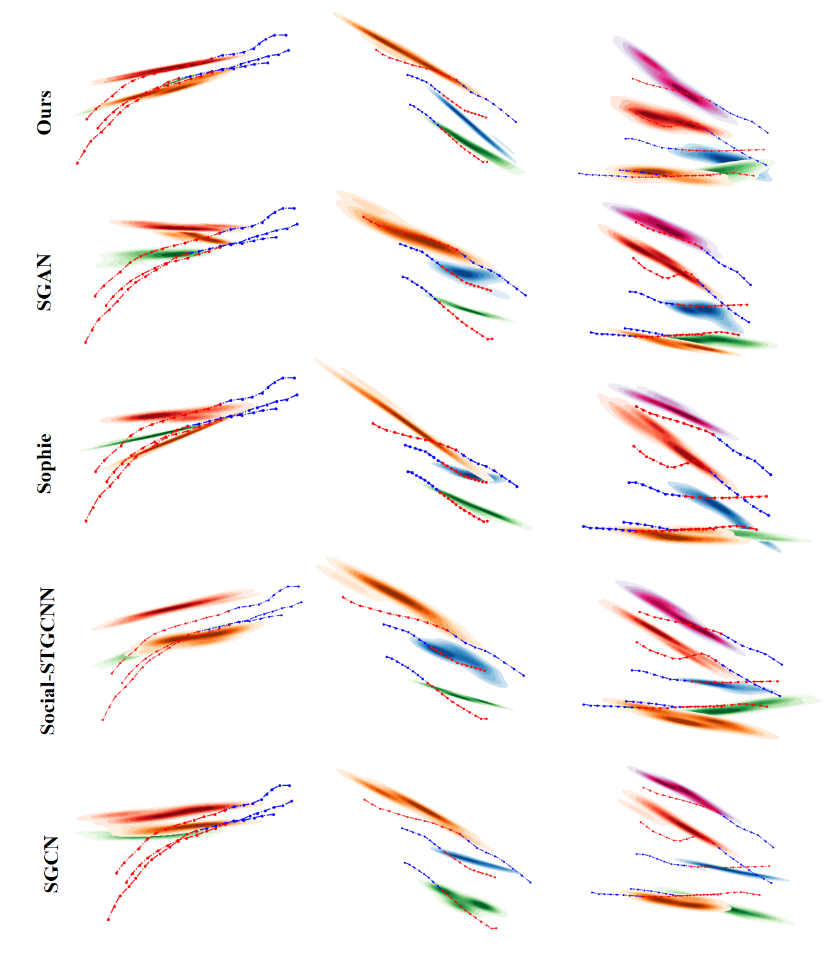}
	\end{center}
	\caption{Visualization of the trajectory distribution predicted by various models in the scenario of pedestrian walking abreast. Different colorful areas represent the future trajectory distribution of different pedestrians. The blue dotted line represents the historical trajectory (8 frames) of the pedestrian, and the red dotted line represents the ground truth (12 frames).}
	\label{fig6}
\end{figure*}

\begin{figure*}[ht]
	\begin{center}
		\includegraphics[width=1.0\linewidth]{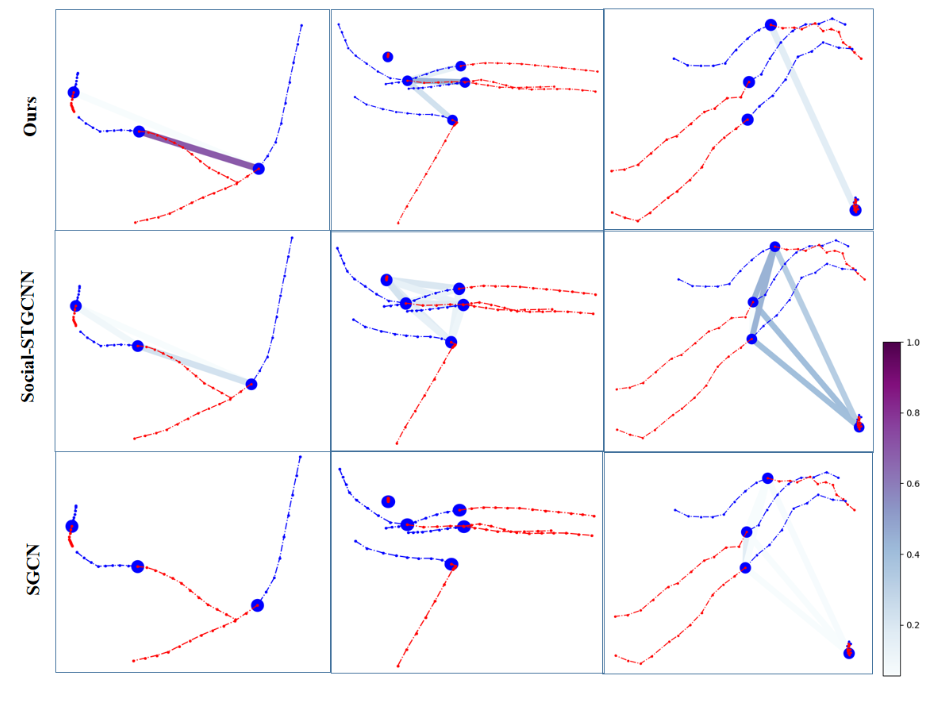}
	\end{center}
	\caption{Visualization of graph relationships in different methods. Each blue node corresponds to one pedestrian and the edge connected to the node itself is not drawn. The blue dotted line represents the historical trajectory (8 frames) of the pedestrian, and the red dotted line represents the ground truth (12 frames). The color of the edge is set according to the colorbar on the right, which describes the weight of the edge. The greater the weight of the edge, the greater the influence between the pedestrians.}
	\label{fig7}
\end{figure*}

The results of quantitative analysis indicate that our model has  high accuracy. Furthermore, we implement qualitative analysis to show the generated trajectories have better social acceptability and physical rationality. As shown in Figure~\ref{fig6} and Figure~\ref{fig7}, we select a number of representative cases from testing datasets.

\par\noindent \textbf{Social acceptability} In Figure~\ref{fig6}, we choose the parallel, encounter, and hybrid cases.
In the first column of Figure~\ref{fig6}, parallel pedestrians generally maintain the same state of motion. The SGAN model built on location-based pooling in this case produces a false prediction of over-avoidance. Attention-based methods, such as Sophie and SGCN, tend to maintain the original state due to self-attention and are insensitive to turning. Social-STGCNN successfully predicts the motion trend but the accuracy is not high. In the second column of Figure~\ref{fig6}, the pedestrian meeting from opposite directions should have some tendency to avoid collisions. But only our method and Social-STGCNN show a tendency to avoid collisions. In the third column of Figure~\ref{fig6}, pedestrians should strike a balance between avoiding collision and maintaining the original state in the mixed case of pedestrian parallelism and encounter. This is a huge challenge for trajectory prediction. It can be seen from Figure~\ref{fig6} that all models fail to accurately predict future trajectories in this complex situation. But our approach produces fewer errors than other methods. The above advantages are attributed to the fact that the SSA function in our method can capture the pedestrian interaction rules more close to the real situation.

In Figure~\ref{fig7}, we visualize the graph relationships of three methods based on weighted matrices and GCN. Our method in the first column does a good job of differentiating and filtering the relationships between agents. Agents with higher collision risk also have larger edge weights. The weights of edges between agents with low collision risk are small or zero. Agents who are walking abreast do not generate edges due to the limitation of cosine in our social soft attention function. In the second column, the filtering effect of the kernel function of Social-STGCNN is so small that the constructed graph is similar to a full connection graph. Moreover, the weight differences between edges in this graph is very small. This makes Social-STGCNN is unable to judge which neighbor has a greater influence on the agent in the face of complex situations. In the third row for SGCN, most edges between adjacent nodes are removed, and only a small number of edges with low weights are left. Due to the influence of self-attention mechanism, in most of the graph relational adjacency matrices established by SGCN, only the diagonal elements have weights. Therefore, the prediction results of SGCN generally tend to maintain the original motion state. From the comparison results in Figure~\ref{fig7}, it can be seen that the graph established by our method is more reasonable and more capable of distinguishing and filtering social interactions.
\begin{figure*}[ht]
	\begin{center}
		\includegraphics[width=1.0\linewidth]{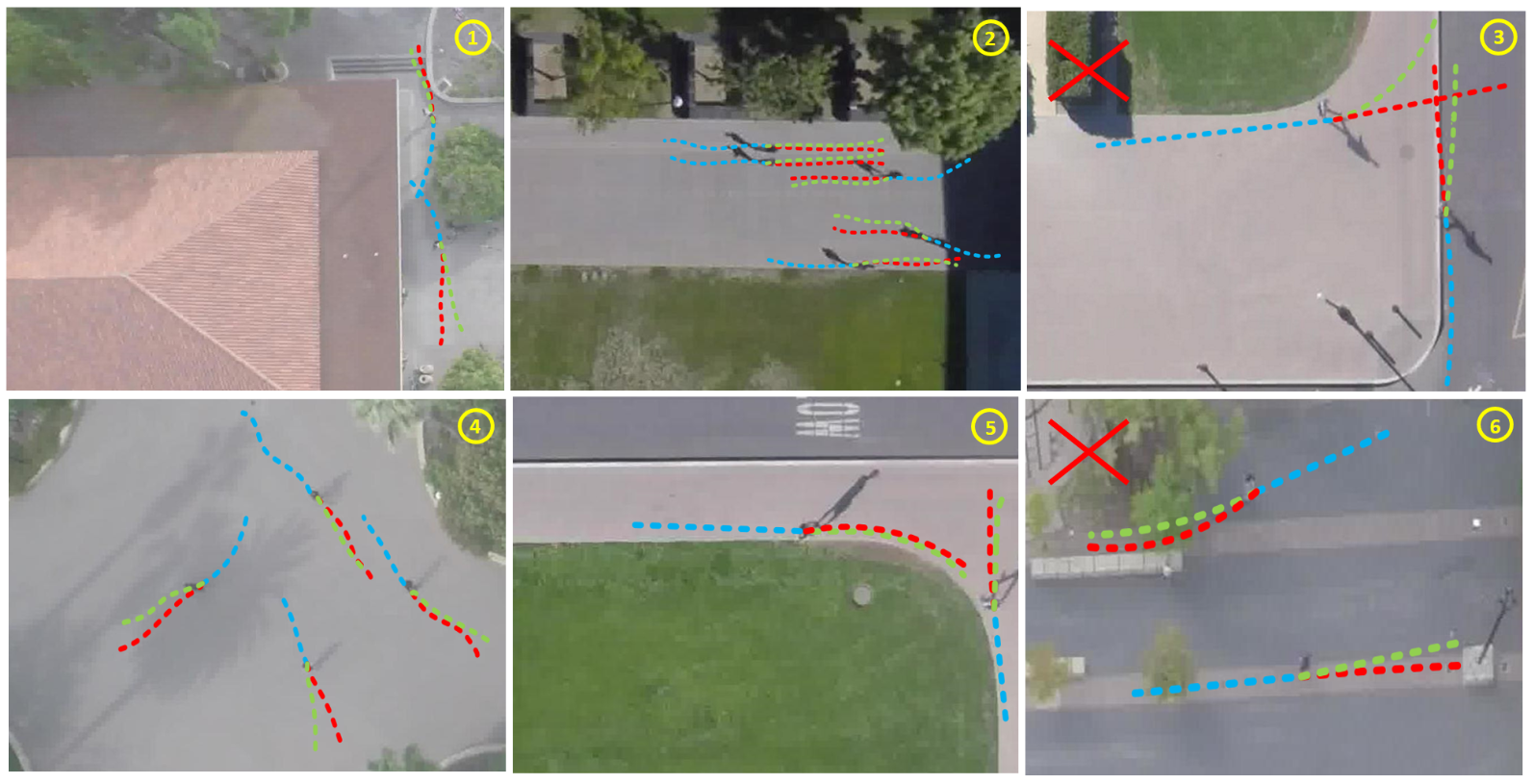}
	\end{center}
	\caption{Visualization of the trajectory prediction of our method on SDD. The green dotted line is the single trajectory predicted by our method, the red dotted line is the ground truth (12 frames), and the blue dotted line is the historical trajectory (8 frames).The unmarked pedestrians in this figure are eliminated in the data processing.}
	\label{fig8}
\end{figure*}

\par\noindent \textbf{Physical rationality} It is widely known that reasonable predictions should conform to realistic physical rules. In Figure~\ref{fig8}, we visualize the predicted results on the SDD dataset to show the model's response to the scene. The SDD contains a large number of roads and obstacles and there are relatively few pedestrians. Therefore, the input of scene image plays a more important role for trajectory prediction on SDD. Our model takes as the input scene images in the form of binary semantics and calculates scene attention at each moment, therefore the whole trajectory in our prediction results will be affected by the scene. In addition, the scene sharing mechanism in our model enables the influence of the scene not to be limited to local areas. For example 2 in Figure~\ref{fig8}, the restriction of the road is transmitted by the pedestrian at the road boundary to the pedestrian in the middle of the road, thus keeping the pedestrian in the middle of the road at a distance from the boundary. In the third column of Figure~\ref{fig8}, examples 3 and 6 show the failure cases. In example 3, the two pedestrians met directly and did not dodge to keep their distance, while in our prediction, the two pedestrians should turn to avoid the collision. In example 6, a pedestrian walked directly towards the obstacle and then stopped to rest, while our prediction concludes that he should bypass the obstacle. Even in this challenging case, the trajectory predicted by our method remains consistent with the rules of physics. We predict that two people who are meeting will follow the road and avoid collisions, while a person walking straight will steer clear of obstacles. This shows that our model can produce reasonable prediction trajectories in most cases.

\section{Conclusion}

In this paper, we propose SSAGCN, a trajectory prediction model, which comprehensively considers a variety of social factors and sequential scene information to obtain accurate and reasonable prediction trajectories. When dealing with the social interaction of pedestrians, we use coordinate data to calculate the relationship of pedestrian speed, direction and distance, and then use social soft attention function to calculate the influences between pedestrians. When considering the context of the scene, we realize that the scene information should influence the trajectory at each moment, and the scene interaction should occur simultaneously with social interaction. Therefore, we calculate the attention at each moment in the scene, and then embed the sequential scene attention attention into the social graph so that the influence of the scene is shared and spread in the graph according to the social relations. Experimental and evaluation results on public data sets show that our method achieves the best results on most datasets due to the new social soft attention function and the scene attention sharing mechanism. Qualitative analysis proves that our model can predict trajectories that are socially and physically acceptable to all pedestrians. In our method, the social soft attention function is an extension of the social force. According to the practical rules, the interaction between pedestrian and collision risk is quantified and applied to the distinction and filtering of adjacency relationship in GCN. We show the rationality of social soft attention function in the visualization of graph relation.
\par Our prediction model can predict reasonable prediction results in various situations, so it can be used for trajectory prediction in complex scenes and provide a reference for collision avoidance for autonomous driving platforms. In our efficiency comparison experiment, we can see that for most prediction methods, their efficiency is not so high in the face of complex and dense scenes. In the future, we will focus on improving the prediction efficiency and accuracy of prediction methods in the scene with dense traffic. A promising improvement is to use sub-graph modeling of social relationships in complex scenarios to reduce computational redundancy and improve efficiency and accuracy.


%





\ifCLASSOPTIONcaptionsoff
  \newpage
\fi



%


{\small
\bibliographystyle{IEEEtran}
\bibliography{IEEEfull}

\begin{thebibliography}{10}
\providecommand{\url}[1]{#1}
\csname url@samestyle\endcsname
\providecommand{\newblock}{\relax}
\providecommand{\bibinfo}[2]{#2}
\providecommand{\BIBentrySTDinterwordspacing}{\spaceskip=0pt\relax}
\providecommand{\BIBentryALTinterwordstretchfactor}{4}
\providecommand{\BIBentryALTinterwordspacing}{\spaceskip=\fontdimen2\font plus
\BIBentryALTinterwordstretchfactor\fontdimen3\font minus
  \fontdimen4\font\relax}
\providecommand{\BIBforeignlanguage}[2]{{%
\expandafter\ifx\csname l@#1\endcsname\relax
\typeout{** WARNING: IEEEtran.bst: No hyphenation pattern has been}%
\typeout{** loaded for the language `#1'. Using the pattern for}%
\typeout{** the default language instead.}%
\else
\language=\csname l@#1\endcsname
\fi
#2}}
\providecommand{\BIBdecl}{\relax}
\BIBdecl

\bibitem{A1}
H.~Bai, S.~Cai, N.~Ye, D.~Hsu, and W.~S. Lee, ``Intention-aware online pomdp
  planning for autonomous driving in a crowd,'' in \emph{2015 ieee
  international conference on robotics and automation (icra)}.\hskip 1em plus
  0.5em minus 0.4em\relax IEEE, 2015, pp. 454--460.

\bibitem{A2}
Y.~Luo, P.~Cai, A.~Bera, D.~Hsu, W.~S. Lee, and D.~Manocha, ``Porca: Modeling
  and planning for autonomous driving among many pedestrians,'' \emph{IEEE
  Robotics and Automation Letters}, vol.~3, no.~4, pp. 3418--3425, 2018.

\bibitem{A3}
P.~Raksincharoensak, T.~Hasegawa, and M.~Nagai, ``Motion planning and control
  of autonomous driving intelligence system based on risk potential
  optimization framework,'' \emph{International Journal of Automotive
  Engineering}, vol.~7, no. AVEC14, pp. 53--60, 2016.

\bibitem{A4}
M.~Luber, J.~A. Stork, G.~D. Tipaldi, and K.~O. Arras, ``People tracking with
  human motion predictions from social forces,'' in \emph{2010 IEEE
  International Conference on Robotics and Automation}.\hskip 1em plus 0.5em
  minus 0.4em\relax IEEE, 2010, pp. 464--469.

\bibitem{A5}
M.~Yasuno, N.~Yasuda, and M.~Aoki, ``Pedestrian detection and tracking in far
  infrared images,'' in \emph{2004 conference on computer vision and pattern
  recognition workshop}.\hskip 1em plus 0.5em minus 0.4em\relax IEEE, 2004, pp.
  125--125.

\bibitem{A6}
B.~Musleh, F.~Garc{\'\i}a, J.~Otamendi, J.~M. Armingol, and A.~De~la Escalera,
  ``Identifying and tracking pedestrians based on sensor fusion and motion
  stability predictions,'' \emph{Sensors}, vol.~10, no.~9, pp. 8028--8053,
  2010.

\bibitem{B1}
H.~Bai, S.~Cai, N.~Ye, D.~Hsu, and W.~S. Lee, ``Intention-aware online pomdp
  planning for autonomous driving in a crowd,'' in \emph{2015 ieee
  international conference on robotics and automation (icra)}.\hskip 1em plus
  0.5em minus 0.4em\relax IEEE, 2015, pp. 454--460.

\bibitem{B2}
J.~Li, H.~Ma, and M.~Tomizuka, ``Conditional generative neural system for
  probabilistic trajectory prediction,'' \emph{arXiv preprint
  arXiv:1905.01631}, 2019.

\bibitem{B3}
A.~Alahi, K.~Goel, V.~Ramanathan, A.~Robicquet, L.~Fei-Fei, and S.~Savarese,
  ``Social lstm: Human trajectory prediction in crowded spaces,'' in
  \emph{Proceedings of the IEEE conference on computer vision and pattern
  recognition}, 2016, pp. 961--971.

\bibitem{B4}
H.~Manh and G.~Alaghband, ``Scene-lstm: A model for human trajectory
  prediction,'' \emph{arXiv preprint arXiv:1808.04018}, 2018.

\bibitem{B5}
J.~Liang, L.~Jiang, J.~C. Niebles, A.~G. Hauptmann, and L.~Fei-Fei, ``Peeking
  into the future: Predicting future person activities and locations in
  videos,'' in \emph{Proceedings of the IEEE Conference on Computer Vision and
  Pattern Recognition}, 2019, pp. 5725--5734.

\bibitem{B6}
P.~Zhang, W.~Ouyang, P.~Zhang, J.~Xue, and N.~Zheng, ``Sr-lstm: State
  refinement for lstm towards pedestrian trajectory prediction,'' in
  \emph{Proceedings of the IEEE Conference on Computer Vision and Pattern
  Recognition}, 2019, pp. 12\,085--12\,094.

\bibitem{C8}
D.~Bahdanau, K.~Cho, and Y.~Bengio, ``Neural machine translation by jointly
  learning to align and translate,'' \emph{arXiv preprint arXiv:1409.0473},
  2014.

\bibitem{B8}
A.~Sadeghian, V.~Kosaraju, A.~Sadeghian, N.~Hirose, H.~Rezatofighi, and
  S.~Savarese, ``Sophie: An attentive gan for predicting paths compliant to
  social and physical constraints,'' in \emph{Proceedings of the IEEE
  Conference on Computer Vision and Pattern Recognition}, 2019, pp. 1349--1358.

\bibitem{B9}
S.~Haddad, M.~Wu, H.~Wei, and S.~K. Lam, ``Situation-aware pedestrian
  trajectory prediction with spatio-temporal attention model,'' \emph{arXiv
  preprint arXiv:1902.05437}, 2019.

\bibitem{B10}
J.~Amirian, J.-B. Hayet, and J.~Pettr{\'e}, ``Social ways: Learning multi-modal
  distributions of pedestrian trajectories with gans,'' in \emph{Proceedings of
  the IEEE Conference on Computer Vision and Pattern Recognition Workshops},
  2019, pp. 0--0.

\bibitem{B15}
Y.~Huang, H.~Bi, Z.~Li, T.~Mao, and Z.~Wang, ``Stgat: Modeling spatial-temporal
  interactions for human trajectory prediction,'' in \emph{Proceedings of the
  IEEE International Conference on Computer Vision}, 2019, pp. 6272--6281.

\bibitem{A16}
V.~Kosaraju, A.~Sadeghian, R.~Mart{\'\i}n-Mart{\'\i}n, I.~Reid, H.~Rezatofighi,
  and S.~Savarese, ``Social-bigat: Multimodal trajectory forecasting using
  bicycle-gan and graph attention networks,'' in \emph{Advances in Neural
  Information Processing Systems}, 2019, pp. 137--146.

\bibitem{C1}
A.~Mohamed, K.~Qian, M.~Elhoseiny, and C.~Claudel, ``Social-stgcnn: A social
  spatio-temporal graph convolutional neural network for human trajectory
  prediction,'' in \emph{Proceedings of the IEEE/CVF Conference on Computer
  Vision and Pattern Recognition}, 2020, pp. 14\,424--14\,432.

\bibitem{E9}
J.~Sekhon and C.~Fleming, ``Scan: A spatial context attentive network for joint
  multi-agent intent prediction,'' \emph{arXiv preprint arXiv:2102.00109},
  2021.

\bibitem{C6}
H.~Sun, Z.~Zhao, and Z.~He, ``Reciprocal learning networks for human trajectory
  prediction,'' in \emph{Proceedings of the IEEE/CVF Conference on Computer
  Vision and Pattern Recognition}, 2020, pp. 7416--7425.

\bibitem{C7}
F.~Marchetti, F.~Becattini, L.~Seidenari, and A.~D. Bimbo, ``Mantra: Memory
  augmented networks for multiple trajectory prediction,'' in \emph{Proceedings
  of the IEEE/CVF Conference on Computer Vision and Pattern Recognition}, 2020,
  pp. 7143--7152.

\bibitem{B11}
N.~Lee, W.~Choi, P.~Vernaza, C.~B. Choy, P.~H. Torr, and M.~Chandraker,
  ``Desire: Distant future prediction in dynamic scenes with interacting
  agents,'' in \emph{Proceedings of the IEEE Conference on Computer Vision and
  Pattern Recognition}, 2017, pp. 336--345.

\bibitem{B12}
H.~Xue, D.~Q. Huynh, and M.~Reynolds, ``Ss-lstm: A hierarchical lstm model for
  pedestrian trajectory prediction,'' in \emph{2018 IEEE Winter Conference on
  Applications of Computer Vision (WACV)}.\hskip 1em plus 0.5em minus
  0.4em\relax IEEE, 2018, pp. 1186--1194.

\bibitem{B13}
F.~Bartoli, G.~Lisanti, L.~Ballan, and A.~Del~Bimbo, ``Context-aware trajectory
  prediction,'' in \emph{2018 24th International Conference on Pattern
  Recognition (ICPR)}.\hskip 1em plus 0.5em minus 0.4em\relax IEEE, 2018, pp.
  1941--1946.

\bibitem{B20}
V.~Kosaraju, A.~Sadeghian, R.~Mart{\'\i}n-Mart{\'\i}n, I.~Reid, H.~Rezatofighi,
  and S.~Savarese, ``Social-bigat: Multimodal trajectory forecasting using
  bicycle-gan and graph attention networks,'' in \emph{Advances in Neural
  Information Processing Systems}, 2019, pp. 137--146.

\bibitem{B18}
S.~Yan, Y.~Xiong, and D.~Lin, ``Spatial temporal graph convolutional networks
  for skeleton-based action recognition,'' \emph{arXiv preprint
  arXiv:1801.07455}, 2018.

\bibitem{E15}
S.~Bai, J.~Z. Kolter, and V.~Koltun, ``An empirical evaluation of generic
  convolutional and recurrent networks for sequence modeling,'' \emph{arXiv
  preprint arXiv:1803.01271}, 2018.

\bibitem{A12}
D.~Helbing and P.~Molnar, ``Social force model for pedestrian dynamics,''
  \emph{Physical review E}, vol.~51, no.~5, p. 4282, 1995.

\bibitem{A13}
S.~Yi, H.~Li, and X.~Wang, ``Understanding pedestrian behaviors from stationary
  crowd groups,'' in \emph{Proceedings of the IEEE Conference on Computer
  Vision and Pattern Recognition}, 2015, pp. 3488--3496.

\bibitem{A9}
B.~T. Morris and M.~M. Trivedi, ``Trajectory learning for activity
  understanding: Unsupervised, multilevel, and long-term adaptive approach,''
  \emph{IEEE transactions on pattern analysis and machine intelligence},
  vol.~33, no.~11, pp. 2287--2301, 2011.

\bibitem{A8}
A.~Sadeghian, F.~Legros, M.~Voisin, R.~Vesel, A.~Alahi, and S.~Savarese,
  ``Car-net: Clairvoyant attentive recurrent network,'' in \emph{Proceedings of
  the European Conference on Computer Vision (ECCV)}, 2018, pp. 151--167.

\bibitem{A15}
L.~Fang, Q.~Jiang, J.~Shi, and B.~Zhou, ``Tpnet: Trajectory proposal network
  for motion prediction,'' in \emph{Proceedings of the IEEE/CVF Conference on
  Computer Vision and Pattern Recognition}, 2020, pp. 6797--6806.

\bibitem{E22}
K.~M. Kitani, B.~D. Ziebart, J.~A. Bagnell, and M.~Hebert, ``Activity
  forecasting,'' in \emph{Computer Vision -- ECCV 2012}, A.~Fitzgibbon,
  S.~Lazebnik, P.~Perona, Y.~Sato, and C.~Schmid, Eds.\hskip 1em plus 0.5em
  minus 0.4em\relax Berlin, Heidelberg: Springer Berlin Heidelberg, 2012, pp.
  201--214.

\bibitem{E1}
Z.~Wu, S.~Pan, F.~Chen, G.~Long, C.~Zhang, and S.~Y. Philip, ``A comprehensive
  survey on graph neural networks,'' \emph{IEEE Transactions on Neural Networks
  and Learning Systems}, 2020.

\bibitem{E2}
M.~Henaff, J.~Bruna, and Y.~LeCun, ``Deep convolutional networks on
  graph-structured data,'' \emph{arXiv preprint arXiv:1506.05163}, 2015.

\bibitem{E3}
J.~Gao, T.~Zhang, and C.~Xu, ``Graph convolutional tracking,'' in
  \emph{Proceedings of the IEEE conference on computer vision and pattern
  recognition}, 2019, pp. 4649--4659.

\bibitem{B21}
J.~Sun, Q.~Jiang, and C.~Lu, ``Recursive social behavior graph for trajectory
  prediction,'' in \emph{Proceedings of the IEEE/CVF Conference on Computer
  Vision and Pattern Recognition}, 2020, pp. 660--669.

\bibitem{A17}
S.~Yan, Y.~Xiong, and D.~Lin, ``Spatial temporal graph convolutional networks
  for skeleton-based action recognition,'' \emph{arXiv preprint
  arXiv:1801.07455}, 2018.

\bibitem{E14}
P.~Veli{\v{c}}kovi{\'c}, G.~Cucurull, A.~Casanova, A.~Romero, P.~Lio, and
  Y.~Bengio, ``Graph attention networks,'' \emph{arXiv preprint
  arXiv:1710.10903}, 2017.

\bibitem{E16}
L.~Shi, L.~Wang, C.~Long, S.~Zhou, M.~Zhou, Z.~Niu, and G.~Hua, ``Sgcn: Sparse
  graph convolution network for pedestrian trajectory prediction,'' in
  \emph{Proceedings of the IEEE/CVF Conference on Computer Vision and Pattern
  Recognition}, 2021, pp. 8994--9003.

\bibitem{A10}
A.~Gupta, J.~Johnson, L.~Fei-Fei, S.~Savarese, and A.~Alahi, ``Social gan:
  Socially acceptable trajectories with generative adversarial networks,'' in
  \emph{Proceedings of the IEEE Conference on Computer Vision and Pattern
  Recognition}, 2018, pp. 2255--2264.

\bibitem{D1}
T.~N. Kipf and M.~Welling, ``Semi-supervised classification with graph
  convolutional networks,'' \emph{arXiv preprint arXiv:1609.02907}, 2016.

\bibitem{A19}
S.~Pellegrini, A.~Ess, and L.~Van~Gool, ``Improving data association by joint
  modeling of pedestrian trajectories and groupings,'' in \emph{European
  conference on computer vision}.\hskip 1em plus 0.5em minus 0.4em\relax
  Springer, 2010, pp. 452--465.

\bibitem{A20}
A.~Lerner, Y.~Chrysanthou, and D.~Lischinski, ``Crowds by example,'' in
  \emph{Computer graphics forum}, vol.~26.\hskip 1em plus 0.5em minus
  0.4em\relax Wiley Online Library, 2007, pp. 655--664.

\bibitem{E12}
A.~Robicquet, A.~Sadeghian, A.~Alahi, and S.~Savarese, ``Learning social
  etiquette: Human trajectory understanding in crowded scenes,'' in
  \emph{European conference on computer vision}.\hskip 1em plus 0.5em minus
  0.4em\relax Springer, 2016, pp. 549--565.

\bibitem{E4}
Y.~Hu, S.~Chen, Y.~Zhang, and X.~Gu, ``Collaborative motion prediction via
  neural motion message passing,'' in \emph{Proceedings of the IEEE/CVF
  Conference on Computer Vision and Pattern Recognition}, 2020, pp. 6319--6328.

\bibitem{E8}
C.~Yu, X.~Ma, J.~Ren, H.~Zhao, and S.~Yi, ``Spatio-temporal graph transformer
  networks for pedestrian trajectory prediction,'' in \emph{European Conference
  on Computer Vision}.\hskip 1em plus 0.5em minus 0.4em\relax Springer, 2020,
  pp. 507--523.

\bibitem{E++}
T.~Salzmann, B.~Ivanovic, P.~Chakravarty, and M.~Pavone, ``Trajectron++:
  Dynamically-feasible trajectory forecasting with heterogeneous data,'' in
  \emph{Computer Vision--ECCV 2020: 16th European Conference, Glasgow, UK,
  August 23--28, 2020, Proceedings, Part XVIII 16}.\hskip 1em plus 0.5em minus
  0.4em\relax Springer, 2020, pp. 683--700.

\bibitem{E10}
M.~Mendieta and H.~Tabkhi, ``Carpe posterum: A convolutional approach for
  real-time pedestrian path prediction,'' \emph{arXiv preprint
  arXiv:2005.12469}, 2020.

\bibitem{E6}
K.~Mangalam, H.~Girase, S.~Agarwal, K.-H. Lee, E.~Adeli, J.~Malik, and
  A.~Gaidon, ``It is not the journey but the destination: Endpoint conditioned
  trajectory prediction,'' in \emph{European Conference on Computer
  Vision}.\hskip 1em plus 0.5em minus 0.4em\relax Springer, 2020, pp. 759--776.

\bibitem{E17}
B.~Yang, G.~Yan, P.~Wang, C.-Y. Chan, X.~Song, and Y.~Chen, ``A novel
  graph-based trajectory predictor with pseudo-oracle,'' \emph{IEEE
  transactions on neural networks and learning systems}, 2021.

\bibitem{E18}
N.~Shafiee, T.~Padir, and E.~Elhamifar, ``Introvert: Human trajectory
  prediction via conditional 3d attention,'' in \emph{Proceedings of the
  IEEE/CVF Conference on Computer Vision and Pattern Recognition (CVPR)}, June
  2021, pp. 16\,815--16\,825.

\bibitem{E19}
B.~Pang, T.~Zhao, X.~Xie, and Y.~N. Wu, ``Trajectory prediction with latent
  belief energy-based model,'' in \emph{Proceedings of the IEEE/CVF Conference
  on Computer Vision and Pattern Recognition (CVPR)}, June 2021, pp.
  11\,814--11\,824.

\bibitem{E21}
K.~Mangalam, Y.~An, H.~Girase, and J.~Malik, ``From goals, waypoints \& paths
  to long term human trajectory forecasting,'' in \emph{Proc. International
  Conference on Computer Vision (ICCV)}, Oct. 2021.

\end{thebibliography}
}

%

\begin{IEEEbiography}[{\includegraphics[width=1in,height=1.25in,clip,keepaspectratio]{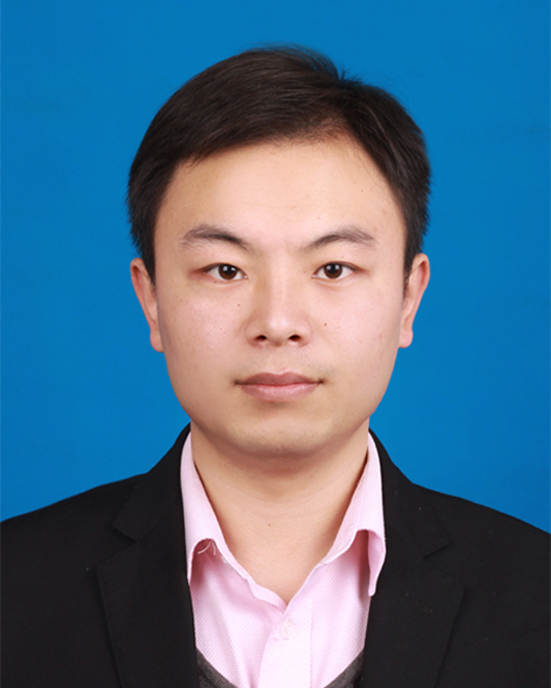}}]{Pei Lv}
received the Ph.D. degree from the State Key Laboratory of CAD\&CG, Zhejiang University Hangzhou, China, in 2013. He is an Associate Professor with the School of Computer and Artificial Intelligence, Zhengzhou University, Zhengzhou, China. His research interests include computer vision and computer graphics. He has authored more than 30 journal and conference papers in the above areas, including the IEEE T RANSACTIONS ON I MAGE P ROCESSING, the IEEE T RANSACTIONS ON C IRCUITS AND SYSTEMS FOR VIDEO TECHNOLOGY, CVPR, ACM MM, and IJCAI.
\end{IEEEbiography}

\begin{IEEEbiography}[{\includegraphics[width=1.25in,height=1.25in,clip,keepaspectratio]{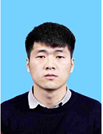}}]{Wentong Wang}received the B.S. degree in Computer Science and Technology from Shenyang Ligong University, Shenyang, China. He is currently pursuing the master's degrees in Henan Institute of Advanced Technology, Zhengzhou University, Zhengzhou, China. His current research is focused on deep learning algorithms and applications for autonomous driving, including driving behavior learning,  pedestrian trajectory prediction and pedestrian intention prediction.
\end{IEEEbiography}

\begin{IEEEbiography}[{\includegraphics[width=1in,height=1.25in,clip,keepaspectratio]{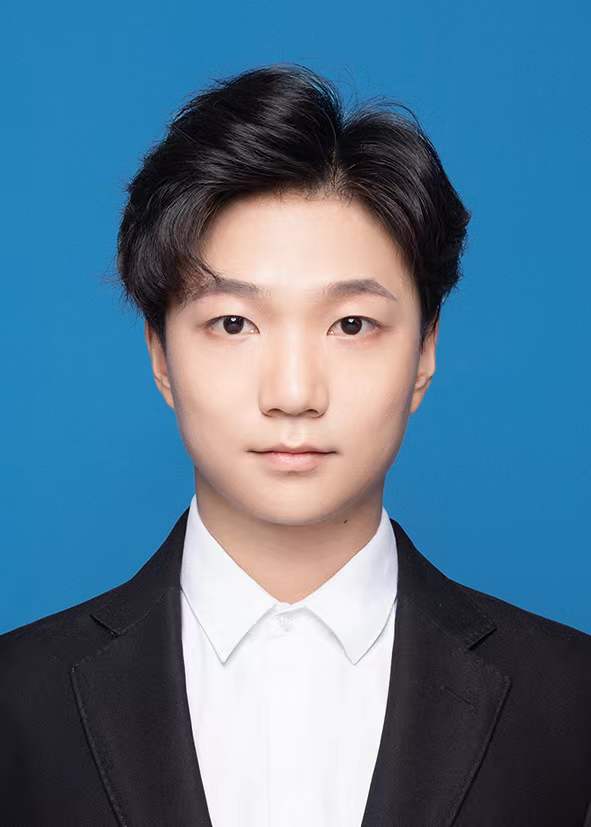}}]{Yunxin Wang}
received the B.S. degree information and Computing Sciences from Northeast Electric Power University,Jilin,China. He is currently pursuing the master's degrees in Henan Institute of Advanced Technology, Zhengzhou University, Zhengzhou, China. His research interests mainly include graph convolutional networks ,pedestrian trajectory prediction and action recognition.
\end{IEEEbiography}

\begin{IEEEbiography}[{\includegraphics[width=1in,height=1.25in,clip,keepaspectratio]{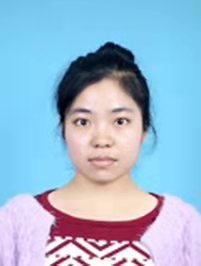}}]{Yuzhen Zhang}
received the B.S. and master's degrees in software engineering from Henan Polytechnic University, Jiaozuo, China. She is currently pursuing the Ph.D. degree in School of Computer and Artificial Intelligence, Zhengzhou University, Zhengzhou, China. Her current research interests include machine learning, computer vision and their applications to motion prediction, scene understanding, and interaction modeling for intelligent autonomous systems.
\end{IEEEbiography}

\begin{IEEEbiography}[{\includegraphics[width=1in,height=1.25in,clip,keepaspectratio]{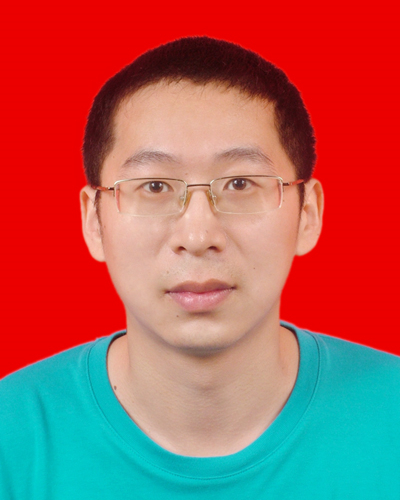}}]{Mingliang Xu}
received the Ph.D. degree in computer science and technology from the State Key Laboratory of CAD\&CG, Zhejiang University, Hangzhou, China, in 2012. He is a Full Professor with the School of Computer and Artificial Intelligence, Zhengzhou University, Zhengzhou, China, where he is currently the Director of the Center for Interdisciplinary Information Science Research and the Vice General Secretary of ACM SIGAI China. His research interests include computer graphics, multimedia, and artificial intelligence. He has authored more than 60 journal and conference papers in the above areas, including the ACM Transactions on Graphics, the ACM Transactions on Intelligent Systems and Technology, the IEEE T RANSACTIONS ON P ATTERN A NALYSIS AND M ACHINE I NTELLIGENCE , the IEEE T RANSACTIONS ON I MAGE P ROCESSING , the IEEE T RANSACTIONS ON C YBERNETICS , the IEEE T RANSACTIONS ON C IRCUITS AND S YSTEMS FOR V IDEO T ECHNOLOGY , ACM SIGGRAPH (Asia), ACM MM, and ICCV.
\end{IEEEbiography}

\begin{IEEEbiography}[{\includegraphics[width=1in,height=1.25in,clip,keepaspectratio]{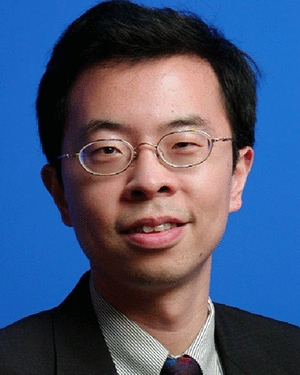}}]{Changsheng Xu}
(Fellow, IEEE) is a Distinguished Professor with the National Laboratory of Pattern Recognition, Institute of Automation, Chinese Academy of Sciences. His research interests include multimedia content analysis/indexing/retrieval, pattern recognition and computer vision. He holds 40 granted/pending patents and published over 300 refereed research papers in these areas. Dr. Xu has served as an Associate Editor, Guest Editor, General Chair, Program Chair, Area/Track Chair, Special Session Organizer, Session Chair, and TPC Member for over 20 IEEE and ACM prestigious multimedia journals, conferences and workshops, including IEEE Transaction on Multimedia, ACM Transaction on Multimedia Computing, Communications and Applications and ACM Multimedia conference. He is an IAPR Fellow and ACM Distinguished Scientist.
\end{IEEEbiography}




\end{document}